%% file: main.tex
\documentclass[11pt]{article}

\usepackage[preprint]{acl}

\usepackage{times}
\usepackage{latexsym}
\usepackage[T1]{fontenc}
\usepackage[utf8]{inputenc}
\usepackage{microtype}
\usepackage{inconsolata}
\usepackage{upquote}
\usepackage{listings}
\usepackage{graphicx}
\usepackage{booktabs}
\usepackage{xcolor}
\usepackage{amsmath}
\usepackage{pgfplots}
\pgfplotsset{compat=1.16}
\usetikzlibrary{positioning}

\title{LuthorSQL: A One-Pass Text-to-SQL Model as System~1\\
for a Slower LLM Agent}

\author{
  \textbf{Ibrahim Memon} \\
  UC Irvine \\
  \texttt{iamemon@uci.edu}
  \And
  \textbf{Armin Karamzade} \\
  UC Irvine \\
  \texttt{karamzaa@uci.edu}
  \AND
  \textbf{M. Hanif Jhaveri} \\
  Independent Researcher \\
  \texttt{hanifj@alumni.stanford.edu}
  \And
  \textbf{Roy Fox} \\
  UC Irvine \\
  \texttt{royf@uci.edu}
}

\begin{document}
\maketitle

\input{sections/abstract}
\input{sections/intro}
\input{sections/related}
\input{sections/method}
\input{sections/system2}
\input{sections/setup}
\input{sections/results}
\input{sections/conclusion}
\input{sections/limitations}

\section*{Acknowledgments}
We thank the BIRD team for evaluating our submission on the hidden test set.

{\sloppy\bibliography{references}}

\appendix
\input{sections/history}
\input{sections/negative}
\input{sections/app_single}
\input{sections/app_valhard}
\input{sections/app_setup}
\input{sections/app_results}
\input{sections/app_tests}

\input{sections/appendix}

\end{document}

%% file: sections/abstract.tex
\begin{abstract}
Frontier LLM agents that query databases with tools are among the most accurate text-to-SQL systems, but per-query calls to a hosted model are slow, costly, and expose sensitive organizational queries to external APIs. For example, a Claude Opus~5.5 agent serving 10,000 queries a day costs over \$70,000 a year at list price. To what extent can local hardware already owned by an organization offload this burden without compromising the agent's accuracy? We present LuthorSQL, a non-autoregressive 7B text-to-SQL model that serves as a rapid System~1 gate for a slower LLM agent (System~2), escalating queries only when uncertain. LuthorSQL makes a single forward pass to predict slot scores over a bounded grammar for query plans, compiles and runs up to 32 candidate plans, and chooses an answer using a trained validator. On its own, LuthorSQL reaches 68.03\% execution accuracy on BIRD's hidden test set at 0.84\,s per question on a single GPU, maintains its accuracy in 9\,GB with 8-bit quantized weights, and transfers zero-shot to Snowflake, BigQuery, MySQL, PostgreSQL, and DuckDB within 3.1 points of its native SQLite accuracy. Combined with Claude Opus~5.5 in a pre-registered confidence cascade over 227 questions on databases published after Opus's training cutoff, LuthorSQL answers nearly half of all questions locally and cuts API cost by 33.3\% at the same accuracy. Finally, contamination probing shows that Opus reproduces 48--92\% of public benchmark gold queries verbatim without seeing the schema; however, its tool-assisted accuracy holds on renamed schemas and reaches 95.6--97.4\% on post-cutoff databases. Code and data: \url{https://github.com/IAMIbrahimmemon/luthorsql}.
\end{abstract}

%% file: sections/intro.tex
\section{Introduction}

\input{figures/fig_example}

Most questions asked of an organization's data are routine: how many orders shipped last month, which customers lapsed, what a region earned.
On BIRD \citep{li2023bird}, a benchmark of large real databases with noisy values, many high-scoring systems write SQL token by token and sample several queries per question before voting, reranking or revising \citep{pourreza2024chasesql,li2025omnisql,yao2025arctic}, and LLM agents that explore the database with tools are among the most accurate.
Run as such an agent, Claude Opus~5.5 answered 77.26\% of 796 questions on held-out BIRD training databases correctly in our runs, at about 2 cents per question at list price. For an organization asking ten thousand questions a day that is about \$200 a day, over \$70,000 a year, with every question sent to a hosted model, even though BIRD hands the agent an expert-written hint per question that real users do not. Yet of the 311 questions it answered correctly in a 400-question sample, our one-pass 7B model alone also got 228 right.

Kahneman's split between a fast, automatic System~1 and a slow, deliberate System~2 \citep{kahneman2011thinking} fits this situation, as do model cascades that send easy inputs to a cheap model \citep{chen2023frugalgpt}, and our design was inspired by TypeSafe's Jev, a ``System One'' model that returns typed decisions instead of free text \citep{almeida2026jev}.
Our System~1 is a one-pass, non-autoregressive 7B model that runs on hardware the organization already owns, where each question it answers costs only electricity and never leaves that machine; System~2, the agent, is called only when System~1 is unsure.
We ask how much of the agent's bill one local model can remove without costing accuracy, whether such a model is tied to the SQL engine it was trained on, and how far an agent's accuracy on public benchmarks can be trusted when the agent may have seen them.

LuthorSQL goes back to slot filling \citep{xu2017sqlnet,yu2018syntaxsqlnet}.
A LoRA adapter \citep{hu2022lora} on OmniSQL-7B \citep{li2025omnisql} reads the schema and question once and fills every slot of a bounded plan grammar in parallel; a deterministic compiler turns each plan into SQL, taking joins from the foreign-key graph; and a second adapter scores the executed candidates and picks one (Figure~\ref{fig:example}).
The grammar covers 84.6\% of BIRD dev gold queries, one reason to pair System~1 with an agent.
Both adapters sit on the same frozen backbone weights, which can be loaded once, so LuthorSQL is one 7B model with two small adapters, not an ensemble of LLMs; LuthorSQL means this two-adapter system throughout (single-adapter variant: Appendix~\ref{sec:single}; path to this design and failed attempts: Appendices~\ref{sec:history} and~\ref{sec:negative}).

\paragraph{Contributions.}
\begin{itemize}\setlength\itemsep{0pt}
\item LuthorSQL, a one-pass slot-filling System~1 on a 7B backbone with a typed plan grammar, lazy $k$-best candidates and a validator trained on hard negatives: 68.03\% on BIRD's hidden test set at 0.84\,s per question on one GPU, the same accuracy with 8-bit weights in about 9\,GB, and, translated without retraining, 57.1--59.3\% on Snowflake, BigQuery, MySQL, PostgreSQL and DuckDB against 60.17\% in SQLite on BIRD dev.
\item A confidence cascade with an LLM agent, tested in registered experiments on data that played no part in choosing it. Calling Opus~5.5 only below a validator margin of 3.0 answered about half of the questions locally and cut the cost per question by 24.8\%, 32.9\% and 32.5\% in three runs on 348 held-out BIRD questions (30.1\% pooled) and by 19.9\% on 693 KaggleDBQA and Spider-Realistic questions compared with Opus alone, at accuracy within 1.2 points of it (none of the differences significant; Figure~\ref{fig:forest}, Table~\ref{tab:tests}). Against Opus called directly through the API, the saving is 33.3\% on 227 questions over unseen databases at equal accuracy.
\item A memorization check for LLM baselines, which we recommend other work run. Without any schema, Opus reproduces 48--92\% of the gold queries of BIRD train, KaggleDBQA and Spider-Realistic verbatim, so thinly prompted LLM scores on these sets partly reflect recall; with every name in the test databases changed, and on 227 new questions over databases published after its training cutoff (97.4\% for Opus alone, 94.3\% for the cascade), its accuracy with a tool holds up (Section~\ref{sec:contam}).
\end{itemize}

%% file: figures/fig_example.tex
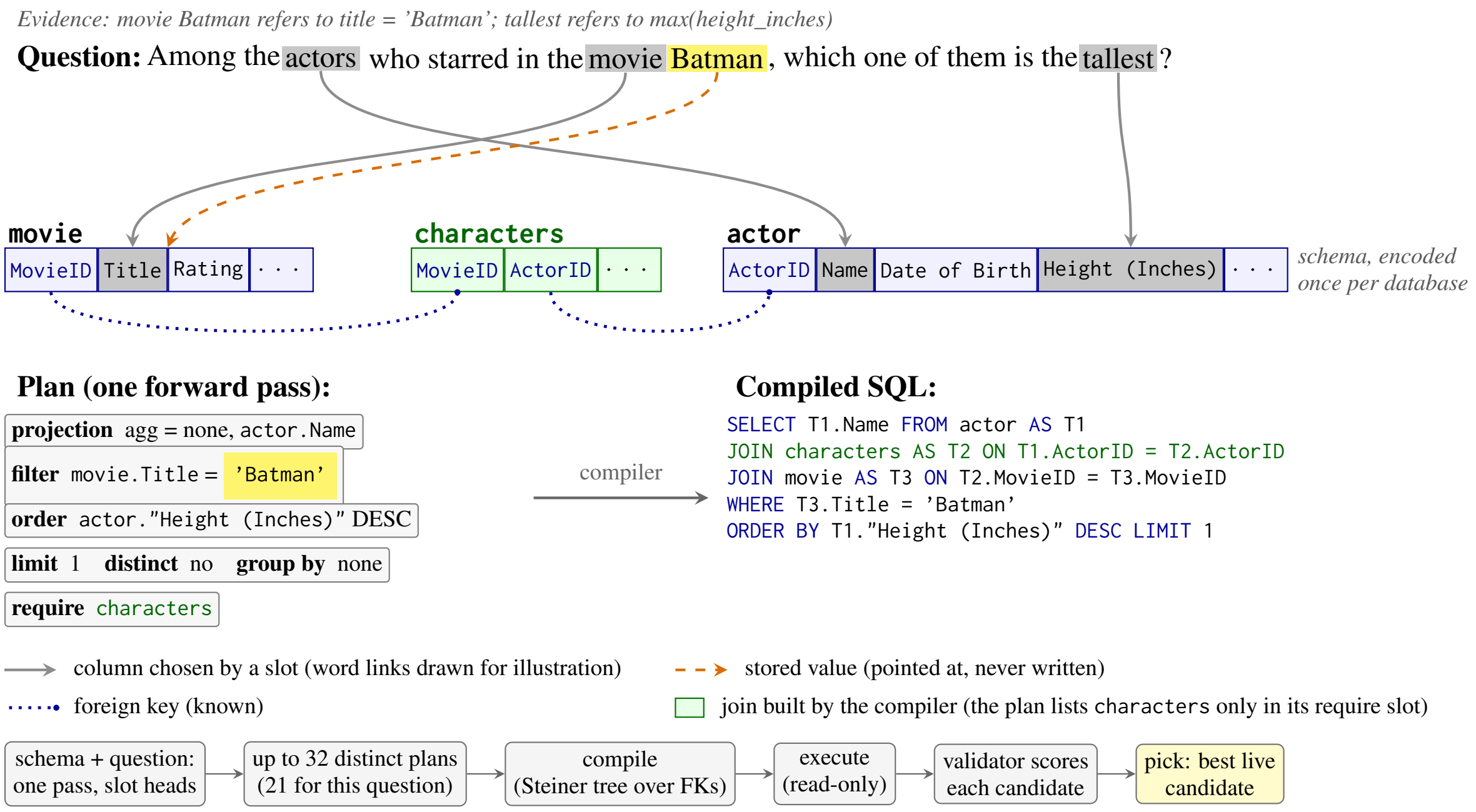
\begin{figure*}[t]
\centering
\begin{tikzpicture}[
  font=\footnotesize,
  col/.style={draw=blue!55!black, fill=blue!6, minimum height=4.6mm, inner sep=1.6pt, font=\ttfamily\scriptsize},
  key/.style={col, font=\ttfamily\scriptsize, text=blue!55!black},
  hit/.style={col, fill=black!22},
  br/.style={fill=green!9, draw=green!45!black},
  tab/.style={font=\ttfamily\bfseries\footnotesize, anchor=south west, inner sep=1pt},
  slotlink/.style={->, >=stealth, draw=black!45, line width=0.8pt},
  vallink/.style={->, >=stealth, draw=orange!85!black, dashed, line width=0.8pt},
  fk/.style={draw=blue!60!black, dotted, line width=0.9pt},
  slot/.style={draw=black!50, fill=black!3, rounded corners=1pt, inner sep=2pt, font=\scriptsize, anchor=west},
  stage/.style={draw=black!55, fill=black!4, rounded corners=2pt, inner sep=2.5pt, font=\scriptsize, align=center},
  hdr/.style={anchor=west, font=\bfseries\small},
]

\node[anchor=west, font=\scriptsize\itshape, text=black!65] at (0,5.15)
  {Evidence: movie Batman refers to title = 'Batman'; tallest refers to max(height\_inches)};
\node[hdr] at (0,4.75) {Question:};
\node[anchor=west, inner sep=0] (q0) at (1.5,4.75) {Among the};
\node[anchor=west, inner sep=1pt, fill=black!22] (qa) at (q0.east) {actors};
\node[anchor=west, inner sep=0] (q1) at (qa.east) {\ who starred in the};
\node[anchor=west, inner sep=1pt, fill=black!22] (qm) at (q1.east) {movie};
\node[anchor=west, inner sep=1pt, fill=yellow!70] (qb) at (qm.east) {Batman};
\node[anchor=west, inner sep=0] (q2) at (qb.east) {, which one of them is the};
\node[anchor=west, inner sep=1pt, fill=black!22] (qt) at (q2.east) {tallest};
\node[anchor=west, inner sep=0] at (qt.east) {?};

\node[tab] at (0,2.75) {movie};
\node[key, anchor=north west] (mid) at (0,2.75) {MovieID};
\node[hit, anchor=west] (mtitle) at (mid.east) {Title};
\node[col, anchor=west] (mrat) at (mtitle.east) {Rating};
\node[col, anchor=west] (mdots) at (mrat.east) {\dots};

\node[tab, text=green!40!black] at (4.3,2.75) {characters};
\node[key, br, anchor=north west] (cmid) at (4.3,2.75) {MovieID};
\node[key, br, anchor=west] (caid) at (cmid.east) {ActorID};
\node[col, br, anchor=west] (cdots) at (caid.east) {\dots};

\node[tab] at (7.6,2.75) {actor};
\node[key, anchor=north west] (aid) at (7.6,2.75) {ActorID};
\node[hit, anchor=west] (aname) at (aid.east) {Name};
\node[col, anchor=west] (abirth) at (aname.east) {Date of Birth};
\node[hit, anchor=west] (aheight) at (abirth.east) {Height (Inches)};
\node[col, anchor=west] (adots) at (aheight.east) {\dots};

\node[anchor=west, font=\scriptsize\itshape, text=black!65, align=left] at (13.55,2.5) {schema, encoded\\once per database};

\draw[fk] (mid.south) .. controls +(0,-0.55) and +(0,-0.55) .. (cmid.south);
\draw[fk] (caid.south) .. controls +(0,-0.55) and +(0,-0.55) .. (aid.south);
\fill[blue!60!black] (cmid.south) circle (0.9pt) (aid.south) circle (0.9pt);

\draw[slotlink] (qa.south) .. controls +(0,-1.0) and +(0,1.1) .. (aname.north);
\draw[slotlink] (qm.south) .. controls +(0,-1.0) and +(0,1.1) .. (mtitle.north);
\draw[vallink] (qb.south) .. controls +(0,-0.9) and +(0.4,1.0) .. (mtitle.north east);
\draw[slotlink] (qt.south) .. controls +(0,-0.9) and +(0,1.0) .. (aheight.north);

\node[hdr] at (0,1.25) {Plan (one forward pass):};
\node[slot] (s1) at (0,0.8) {\textbf{projection}\ \ agg = none,\ \texttt{actor.Name}};
\node[slot] (s2) at (0,0.33) {\textbf{filter}\ \ \texttt{movie.Title} = \colorbox{yellow!70}{\strut\texttt{'Batman'}}};
\node[slot] (s3) at (0,-0.14) {\textbf{order}\ \ \texttt{actor."Height (Inches)"} DESC};
\node[slot] (s4) at (0,-0.61) {\textbf{limit}\ \ 1\quad \textbf{distinct}\ \ no\quad \textbf{group by}\ \ none};
\node[slot] (s5) at (0,-1.08) {\textbf{require}\ \ \textcolor{green!40!black}{\texttt{characters}}};

\node[hdr] at (7.6,1.25) {Compiled SQL:};
\node[anchor=north west, align=left, font=\ttfamily\scriptsize, inner sep=1pt] (sql) at (7.6,1.0) {%
\textcolor{blue!60!black}{SELECT} T1.Name \textcolor{blue!60!black}{FROM} actor \textcolor{blue!60!black}{AS} T1\\
\textcolor{green!40!black}{JOIN characters AS T2 ON T1.ActorID = T2.ActorID}\\
\textcolor{blue!60!black}{JOIN} movie \textcolor{blue!60!black}{AS} T3 \textcolor{blue!60!black}{ON} T2.MovieID = T3.MovieID\\
\textcolor{blue!60!black}{WHERE} T3.Title = 'Batman'\\
\textcolor{blue!60!black}{ORDER BY} T1."Height (Inches)" \textcolor{blue!60!black}{DESC LIMIT} 1};
\draw[->, >=stealth, black!60, line width=0.8pt] (5.6,0.1) -- node[above, font=\scriptsize] {compiler} (7.45,0.1);

\begin{scope}[shift={(0,-1.72)}]
  \draw[slotlink] (0,0) -- (0.55,0); \node[anchor=west, font=\scriptsize] at (0.6,0) {column chosen by a slot (word links drawn for illustration)};
  \draw[vallink] (7.1,0) -- (7.65,0); \node[anchor=west, font=\scriptsize] at (7.7,0) {stored value (pointed at, never written)};
  \draw[fk] (0.05,-0.4) -- (0.55,-0.4); \fill[blue!60!black] (0.55,-0.4) circle (0.9pt);
  \node[anchor=west, font=\scriptsize] at (0.6,-0.4) {foreign key (known)};
  \fill[green!9] (7.1,-0.5) rectangle (7.4,-0.3); \draw[green!45!black] (7.1,-0.5) rectangle (7.4,-0.3);
  \node[anchor=west, font=\scriptsize] at (7.45,-0.4) {join built by the compiler (the plan lists \texttt{characters} only in its require slot)};
\end{scope}

\begin{scope}[shift={(0,-2.82)}]
  \node[stage, anchor=west] (p1) at (0,0) {schema + question:\\one pass, slot heads};
  \node[stage, right=0.4cm of p1] (p2) {up to 32 distinct plans\\(21 for this question)};
  \node[stage, right=0.4cm of p2] (p3) {compile\\(Steiner tree over FKs)};
  \node[stage, right=0.4cm of p3] (p4) {execute\\(read-only)};
  \node[stage, right=0.4cm of p4] (p5) {validator scores\\each candidate};
  \node[stage, right=0.4cm of p5, fill=yellow!25] (p6) {pick: best live\\candidate};
  \foreach \a/\b in {p1/p2,p2/p3,p3/p4,p4/p5,p5/p6} \draw[->, >=stealth, black!60] (\a) -- (\b);
\end{scope}
\end{tikzpicture}
\caption{How LuthorSQL answers a question. The slot heads, read after the question, point to schema columns and to a
stored value; the word links are drawn for illustration, since the model scores slot states, not word alignments. No
question word names \texttt{characters}; the plan lists it only in its require slot, and the compiler builds the join
along the foreign-key path between \texttt{movie} and \texttt{actor}. The same pass yields up to 32 distinct plans (21
for this question), found by best-first search, which are compiled, executed and scored by the validator. Example: BIRD train question~740 (\texttt{movie}, a held-out database), answered
correctly.}
\label{fig:example}
\end{figure*}

%% file: sections/related.tex
\section{Related Work}

\paragraph{Slot filling and parallel generation.}
SQLNet \citep{xu2017sqlnet} filled a fixed sketch with its own prediction module per slot, after Seq2SQL \citep{zhong2017seq2sql} had split the query into classified and decoded parts; both targeted single-table WikiSQL questions.
Multi-table work moved to grammar-based decoders and intermediate representations decoded step by step \citep{yu2018syntaxsqlnet,guo2019irnet}, building on sketch-then-fill decoding \citep{dong2018coarse}.
Non-autoregressive decoders \citep{gu2018nat,ghazvininejad2019maskpredict} and SmBoP \citep{rubin2021smbop} cut the number of sequential steps, and diffusion language models \citep{nie2025llada,ye2025dream,gong2025diffucoder} denoise iteratively, and \citet{ma2026sqld1} evaluate them on text-to-SQL.
Sketch slot filling has also been applied to multi-table Spider queries \citep{choi2021ryansql}, and SemQL and NatSQL derive joins from the foreign-key graph \citep{guo2019irnet,gan2021natsql}.
Our closed slot grammar, with expression templates mined from training data, is filled by a 7B backbone in one forward pass that returns $k$-best plans.

\paragraph{Selection.}
Picking among candidates by execution agreement \citep{wang2023selfconsistency}, a trained verifier or an LLM judge is common \citep{ni2023lever,pourreza2024chasesql}.
Closest to our validator, GradeSQL scores candidate queries with an outcome reward model at test time \citep{tritto2025gradesql}; XiYan-SQL combines several generators and reorganizes their candidates for a selection model \citep{gao2024xiyansql,liu2026xiyansql}, and MSc-SQL has a small model critique several sampled queries together \citep{gorti2025mscsql}.
R3-SQL groups candidates by execution result and ranks the groups \citep{han2026r3sql}, as MBR-exec and CodeT use execution agreement \citep{shi2022mbrexec,chen2023codet}; our validator instead scores each candidate on its own, and we group by result only to compute its margin and to show the agent one entry per result (more in Appendix~\ref{app:related}).
Our candidates come mainly from lazy $k$-best enumeration over factorized decisions \citep{huang2005kbest}; fixed-seed samples from the same slot scores fill any shortfall.

\paragraph{Cascades and fast/slow systems.}
FrugalGPT \citep{chen2023frugalgpt} sends a query through a chain of models and stops when a scorer trusts the answer, AutoMix routes to a larger model when a smaller model's self-verified answer looks wrong \citep{aggarwal2024automix}, and RouteLLM trains routers on preference data to choose between a strong and a weak model \citep{ong2025routellm}; confidence-based deferral is well studied \citep{varshney2022cascading,jitkrittum2023deferral,ding2024hybrid}, including on answer consistency \citep{yue2024cascades}, like our vote-agreement baseline.
For text-to-SQL, \citet{malekpour2024routing} route among LLMs and argue that cascades cannot be used because a weak model's SQL cannot be judged before falling back; our validator margin is such a judgment. ZeroNL2SQL lets small fine-tuned models write an SQL sketch that an LLM completes \citep{fan2024zeronl2sql}, and EllieSQL routes each question to a cheaper or costlier pipeline by its estimated complexity \citep{zhu2025elliesql}.
TypeSafe's Jev, a ``System One'' model that returns typed decisions with probabilities instead of text and that TypeSafe describes as outputting all probabilities in parallel rather than token by token, inspired our System~1 \citep{almeida2026jev}. SwiftSage pairs a small fast model with a slow LLM planner \citep{lin2023swiftsage}.
Our setting differs in that the fast model is a one-pass slot parser whose executed, result-grouped $k$-best list and validator margin serve both as the routing signal and as the agent's starting point.

\paragraph{Evaluation.}
Execution accuracy on a single database can accept a wrong query that happens to match \citep{zhong2020ts}; our scores share this limitation. BIRD's labels have known errors \citep{wretblad2024understanding}, so we report the original dev gold and label BIRD's 2025 corrected gold \citep{birdsql2025dev} separately, and we test the main differences with paired significance tests \citep{dror2018hitchhiker}, clustered by database.
Benchmark contamination can inflate LLM scores \citep{sainz2023nlp}; guided completion of benchmark instances detects it \citep{golchin2024timetravel}, and text-to-SQL accuracy has been shown to drop on data an LLM cannot have seen \citep{ranaldi2024contamination}. We apply such probes to the agent baseline itself and add a renamed-schema test and a post-cutoff evaluation set.

%% file: sections/method.tex
\section{System~1: LuthorSQL}
\label{sec:method}

Given a natural-language question $q$, an external knowledge hint (``evidence'') $e$, and a relational database $D$ with schema $S$ and stored contents, the text-to-SQL task is to produce an executable SQL query $y$. Following the BIRD benchmark \citep{li2023bird}, a predicted query $y$ is evaluated by execution accuracy: $y$ is correct if executing it on $D$ yields the exact same set of result rows as the ground-truth query $y^*$, written $\mathrm{exec}(y, D) = \mathrm{exec}(y^*, D)$. Within our fast/slow architecture, System~1's role is to return an accurate candidate query from one generator pass and one validator pass on local hardware while producing a score that Section~\ref{sec:s2} converts into a routing confidence for System~2.

\subsection{Overview}
Instead of writing SQL token by token, LuthorSQL predicts the structured query plan in a single forward pass. From the predicted query plan, LuthorSQL then compiles the executable SQL code. After factorizing the plan score, LuthorSQL enumerates top-$k$ choices of that plan, and chooses the final answer through a trained validator (Figure~\ref{fig:example}). Every step is designed to alleviate different failure modes of small language models:
\begin{itemize}
    \item \textbf{Structured plans:} Constraining output to discrete slot decisions prevents the model from generating syntax errors or hallucinating string literals.
    \item \textbf{Deterministic compiler:} Inferring table joins over the foreign-key graph relieves the neural network from writing complex join clauses.
    \item \textbf{Factorized scoring:} Scoring slots independently allows $k$-best candidate plans to be enumerated without autoregressive beam search.
    \item \textbf{Full-query validator:} Scoring the compiled SQL string allows a second adapter to inspect the complete query structure and select among executed candidates.
\end{itemize}

\subsection{Plans and Grammar Coverage}
A query plan $p$ is a bounded record of discrete choices that abstracts away surface SQL syntax while capturing the semantic intent of a query. Formally, each slot $\sigma$ represents a single structural decision over a finite option set $O_\sigma(q, D)$, where options correspond to schema items in $S$, small closed label sets, or grounded literal candidates. A complete plan $p$ assigns one choice to each active slot, and our plan grammar $\mathcal{G}$ defines the space of all valid combinations of slot choices.

To ensure the model never invents filter values, filter slots can only point to a finite set of literal candidates $L(q, D)$ extracted from three sources: (1) database values matched by a text span in $q$ or $e$; (2) numbers, dates, quoted strings, and capitalized phrases parsed directly from $q$ and $e$; and (3) the complete value lists of low-cardinality columns containing at most 12 distinct values. A plan holds one to four projections---each pairing a schema column or row count with an aggregation operator (\texttt{none}, \texttt{count}, \texttt{count distinct}, \texttt{sum}, \texttt{avg}, \texttt{min}, or \texttt{max}) and an optional expression template---alongside up to four filters, each specifying a column, an operator, and a pointer into $L(q, D)$. Additional slots govern grouping, ordering, limits, \texttt{DISTINCT}, date parts, ratios, \texttt{OR} groups, and a single level of derived tables (full slot specification and mined template counts in Appendix~\ref{app:slots}).

We define the coverage of $\mathcal{G}$ as the proportion of benchmark queries with at least one plan $p \in \mathcal{G}$ that when compiled becomes a query that returns $\mathrm{exec}(y^*, D)$. We observe 1,297/1,534 (84.6\%) of the BIRD dev queries and 670/796 (84.2\%) of the DECISION queries covered by converting the gold queries to plans and testing them by execution. An oracle slot model would return 1,280 (83.4\%) of the dev responses. The remaining queries involve novel expression templates, ungroundable literal values, CTEs, and scalar subqueries. To address those queries we are using a cascade with an unconstrained LLM agent.

\subsection{Scoring Plans in One Forward Pass}
Our generator scores every slot in a single forward pass so that the log-probability of any complete plan decomposes as a sum over its individual decisions. We use OmniSQL-7B \citep{li2025omnisql,hui2024qwen25coder} equipped with a LoRA adapter as the encoder backbone. We serialize the database schema $S$ (tables, columns, types, descriptions, and sample values) as a text prefix, followed by $q$, $e$, and one learned embedding token per slot $\sigma$. A single forward pass over this sequence produces a contextualized hidden state $h_\sigma$ for every slot.

To represent the option set $O_\sigma(q, D)$, each schema item is encoded as the mean of the hidden states across its serialized schema line, while each literal candidate in $L(q, D)$ combines the hidden state of its mention span in $q$, the schema state of its host column, and an optional ``Values:'' suffix entry. Specialized lightweight heads then score the decisions: \emph{pointer heads} compare $h_\sigma$ against schema options; \emph{choice heads} apply a softmax over small categorical label sets; \emph{pairwise heads} pass $(h_\sigma, \text{option})$ pairs through a small MLP to score aggregations, operators, templates, and constants conditioned on the chosen column; and \emph{value heads} score literal candidates. Because these decisions factorize, the total score of a plan $p$ is the sum of log-probabilities over the active decisions $d \in p$:
$$\mathrm{score}(p) = \sum_{d \in p} \log P(d \mid q, e, D)$$
We train the generator with cross-entropy against execution-verified gold plans. For set-valued slots such as filters and literal candidates, where multiple target choices $P$ are simultaneously valid against a negative set $N$, we optimize a multi-positive cross-entropy objective in which each positive $i \in P$ competes against all negatives without suppressing other positives:
$$\mathcal{L}_{\text{set}} = -\sum_{i \in P} \log \frac{e^{s_i}}{e^{s_i} + \sum_{j \in N} e^{s_j}}$$

\subsection{From Plan to SQL}
A deterministic compiler translates each plan $p$ into executable SQL and constructs all table joins automatically, ensuring that a join fails only if a model hint is incorrect or the shortest relational path does not match the question's intent. Given the set of tables referenced by a plan's columns, the compiler connects them via a minimum-cost Steiner tree over the database's foreign-key graph, guided by optional model-predicted join hints that specify required bridge tables or preferred foreign keys (declared keys cost 1, inferred keys more and hinted keys less; dead-end bridge tables are pruned). The compiler emits native SQLite syntax and transpiles queries to other enterprise SQL dialects when targeting external database engines.

\subsection{Candidate Enumeration}
Because $\mathrm{score}(p)$ decomposes additively across slots, we can extract the top-$k$ highest-scoring plans directly from the single forward pass without rerunning the backbone. A best-first priority queue pops plans in increasing order of their log-probability gap from the greedy best plan, lazily expanding only the conditional decisions that a plan actively uses. We discard plans that fail to compile or that compile to a duplicate SQL string already in the pool, filling any remaining shortfall via fixed-seed sampling from the slot distributions. This yields up to $k = 32$ distinct SQL queries per question, each executed read-only against $D$ with a 3\,s timeout to form our set of executed \emph{candidates}. We partition these candidates into \emph{result groups}, where each group consists of all candidates that return an identical execution result table.

\subsection{Picking an Answer with the Validator}
Because slot decisions are scored factorially, the generator cannot evaluate global interactions across a complete SQL query; we therefore introduce a validator, \textsc{VALHARD}, to inspect and score each compiled candidate in full. \textsc{VALHARD} is a second LoRA adapter attached to the same frozen OmniSQL-7B backbone with a linear scoring head. For each candidate $c$, it reads the serialized schema $S$, question $q$, evidence $e$, and compiled SQL string, outputting a scalar quality score $s(q, c)$ without access to the candidate's generation rank, plan score, or execution result. We define a candidate as \emph{live} if it executes within the timeout and returns at least one non-\texttt{NULL} value. At inference time, LuthorSQL outputs the highest-scoring live candidate:
$$\hat{y} = \arg\max_{c \text{ live}} s(q, c)$$
falling back to the highest-scoring candidate overall if no candidate is live.

We train \textsc{VALHARD} with the generator's $k = 32$ candidate lists of BIRD-train databases. It is trained using hard negatives covering common ranking errors: near-miss wrong result groups and synthetic query perturbations (e.g., \texttt{COUNT} $\to$ \texttt{COUNT DISTINCT} switching, filter removal, \texttt{ORDER BY} switching) kept only if they return a new, wrong, non-null result on execution. Although the validator scores each candidate on its own, its training loss treats each result group as one composite option. The log-mean-exp of the result group candidates' scores is used to compute the probability mass to avoid overestimating the probability mass for replicated SQL strings (Appendix~\ref{app:valhard}).

%% file: sections/system2.tex
\section{Pairing System~1 with an LLM Agent}
\label{sec:s2}

System~2 is Claude Opus~5.5 run as an agent.\footnote{The agent transcripts of every held-out test and of the BIRD dev run record the model identifier \texttt{claude-opus-5-5}. It was run through Claude Code subagents on a subscription, except the API runs of Appendix~\ref{app:plainapi}; see Appendix~\ref{sec:setup2}.}
Its database tool is a read-only SQLite runner, called through the shell, that can also print the schema (CREATE statements, three sample rows per table and BIRD's column descriptions); as a Claude Code agent it also has file and shell tools, which it used to read its questions and write its answers, and it is instructed to execute at most eight queries before giving one final SQL query.
Alone, this agent scores 77.26\% on our 796-question selection split, against 59.05\% for System~1.
We combine the two in two ways.

\paragraph{Candidate handoff.}
System~1 runs first.
Its candidates (up to 32) are grouped by execution result and the top eight groups, ordered by validator score, go into the agent's prompt as labeled options, each with its SQL, a short preview of its result and how many of its candidates returned it.
The agent keeps the same tool and the eight-query limit; its prompt adds the candidates and asks it to pick a clearly right one without further exploration, or to write its own query only when it is confident every candidate is wrong.
On 126 of the 796 questions, the ones whose gold query the grammar cannot express, System~1 was not run in our DECISION experiments, a choice that uses gold; there the agent gets no candidates (under the handoff prompt) and System~1 is scored as wrong; 70 of the 400 sampled questions are of this kind. In the cascade these 126 and 26 more questions whose candidates form fewer than two result groups have no margin, so 152 questions always go to the agent.
We test whether checking a short list of executed answers takes fewer tool rounds than building a query from the schema up.

\paragraph{Confidence cascade.}
System~1 answers alone when it is confident and passes the question to the handoff setup otherwise.
Confidence is the validator's score margin between its top two result groups, and on DECISION the 126 out-of-grammar questions always go to the agent; the held-out tests route every question by the margin alone, with no gold.
We pick the threshold with a rule written down before the run: on a 0.25 grid from 2.0 to 6.0, take the threshold that maximizes accuracy minus cost, valuing a right answer at \$1 and an agent call at its list price.

Neither setup changes System~1.

%% file: sections/setup.tex
\section{Experimental Setup}
\label{sec:setup}

\paragraph{Data and splits.}
All training data is public: BIRD train \citep{li2023bird}, Spider train \citep{yu2018spider} and a sample of SynSQL-2.5M \citep{li2025omnisql}, each gold query kept only if its compiled plan returns the gold result (details in Appendix~\ref{app:setup}).
BIRD dev (1,534 questions, 11 databases) is scored with the original 2024-06-27 gold, as in most prior work, and we label rescoring against BIRD's 2025 revision \citep{birdsql2025dev} wherever it appears.
From October 8 on, every design choice was made on DECISION, 796 questions from 8 BIRD-train databases that our adapters never trained on; RESERVED holds 348 more questions from the same databases.
OmniSQL's own fine-tuning data includes all of BIRD train, so these databases are held out from our adapters but not from the backbone; Appendix~\ref{app:setup} lists the databases and other limits on how clean the splits are.
We allowed at most three looks at BIRD dev from October 8: LuthorSQL used the first and the System~2 dev run the second.

\paragraph{Metric and tests.}
For queries that lie outside the plan grammar $\mathcal{G}$, System~1 still outputs its highest-scoring valid plan when evaluated on BIRD dev, whereas on our held-out DECISION split out-of-grammar questions are strictly scored as incorrect in all standalone System~1 arms. Latency measurements are reported on a seeded 200-question DECISION sample at batch size 1 (averaging 29.6 distinct candidates per question; full hardware details in Appendix~\ref{app:setup}). Paired comparisons use an exact two-sided sign-flip test over per-database summed differences, with 95\% intervals by inversion; with 8 databases the smallest attainable $p$ is $2/2^8 \approx 0.008$, so we also report question-level McNemar wins and losses.

\paragraph{System~2 runs.}
Each agent run gives one Claude Code subagent 25 questions at a time, with the read-only SQL tool and the prompts in Appendix~\ref{app:prompts}.
The handoff comparison uses a fixed 400-question DECISION sample with the two arms interleaved; the cascade threshold is chosen on all 796 questions after a second handoff run on the other 396.
Costs are Opus~5.5 list prices applied to the subagents' usage logs, not invoices (Appendix~\ref{sec:setup2}).

\paragraph{Held-out tests.}
The cascade was tested on three held-out sets, each registered before any agent ran on it: RESERVED; KaggleDBQA's test split \citep{lee2021kaggledbqa} and Spider-Realistic \citep{deng2021structure}, on 27 databases, two of which overlap System~1's training data (Section~\ref{sec:s2results}); and a new set of 227 questions over 8 databases built from public data dated after the agent's training cutoff (Appendix~\ref{app:fresh}). Memorization probes, a renamed-schema test and attempts to raise accuracy were registered the same way (Appendices~\ref{app:probes} and~\ref{app:tuning}); one later exploratory search was not. Except for the API runs (Appendix~\ref{app:plainapi}), the agent ran as Claude Code subagents in a sandbox holding only the databases and the SQL tool, and every transcript was scanned for access outside it.

%% file: sections/results.tex
\section{Results}
\label{sec:results}

\subsection{System~1 Standalone Performance and Latency}
LuthorSQL achieves competitive accuracy among fast local models while operating in under a second on a single GPU. On BIRD dev, the two-adapter System~1 reaches 59.39\% execution accuracy (911 of 1,534 questions), placing it roughly on par with multi-sample diffusion language models with a reported latency of over five seconds (a preprint result under different conditions and hardware), though behind slower autoregressive models that generate SQL token by token such as its OmniSQL-7B backbone (62.6\% in our harness; Appendix~\ref{app:moreresults}). As shown in Table~\ref{tab:latency} (Appendix~\ref{app:moreresults}), generating only the single greedy plan takes just 0.13\,s end-to-end on one H100 PCIe and achieves 49.12\% on DECISION. Expanding to $k = 32$ candidates and reranking with \textsc{VALHARD} adds nearly ten percentage points of accuracy (+9.93 points) for roughly 0.7\,s of additional compute, yielding a median end-to-end latency of 0.84\,s per question (\$0.64 per 1,000 queries).

\paragraph{Zero-shot dialect transfer.} Because LuthorSQL predicts an engine-agnostic query plan before compiling to SQL, it transfers across enterprise database engines without retraining. Transpiling the SQLite candidates and executing them natively on all 1,534 BIRD dev questions yields 57.1--59.3\% on Snowflake, MySQL, DuckDB, PostgreSQL and BigQuery, compared to 60.17\% in SQLite on the same database files (Table~\ref{tab:dialects} in Appendix~\ref{app:moreresults}). A single local checkpoint thus serves six distinct SQL engines within a 3.1-point accuracy band.

\subsection{Generalization to the BIRD Hidden Test Set}
\label{sec:test}

\begin{table}[t]
\centering\small
\setlength{\tabcolsep}{3pt}
\begin{tabular}{@{}lcccc@{}}
\toprule
 & Simple & Moderate & Challenging & Total \\
\midrule
Test EX & 77.24 & 64.86 & 43.51 & 68.03 \\
Test Soft F1 & 77.79 & 66.74 & 45.77 & 69.26 \\
Test R-VES & 75.04 & 60.85 & 40.69 & 65.17 \\
\midrule
Dev EX & 69.41 & 45.69 & 39.31 & 59.39 \\
Test $-$ dev EX & +7.83 & +19.17 & +4.20 & +8.64 \\
\bottomrule
\end{tabular}
\caption{LuthorSQL (two adapters) on BIRD's hidden test set (1,789 questions: 949 simple, 555 moderate, 285 challenging), run by BIRD, against our single BIRD dev scoring (original gold, 1,534 questions).}
\label{tab:test}
\end{table}
On BIRD's official hidden test set of 1,789 questions, LuthorSQL reaches 68.03\% execution accuracy (69.26\% Soft F1, 65.17\% R-VES; Table~\ref{tab:test}), and our single-adapter variant achieves 68.19\% (Appendix~\ref{sec:single}), matching the published single-model leaderboard score of the 32B Arctic-ExCoT model \citep{zhai2025excot} at a fraction of the parameter count. Our test score is 8.64 points above our dev score, a gap we cannot explain without test labels, so we rely on BIRD dev and our held-out splits for all controlled ablations.

\subsection{Pre-Registered Cascade Evaluations}
\label{sec:s2results}

\begin{table}[t]
\centering\small
\setlength{\tabcolsep}{2pt}
\resizebox{\columnwidth}{!}{\begin{tabular}{@{}lcccc@{}}
\toprule
 & RES run 1 & RES pool & External & Fresh \\
\midrule
Questions / DBs & 348 / 8 & 3$\times$348 / 8 & 693 / 27 & 227 / 8 \\
Agent calls & 50.9\% & 50.9\% & 51.9\% & 55.1\% \\
\midrule
EX Opus (B0) & 80.17 & 79.50 & 88.31 & 97.36 \\
EX Opus + cands (B1) & 80.46 & 80.75 & 86.87 & -- \\
EX cascade (C) & 81.32 & 80.46 & 88.17 & 94.27 \\
EX System~1 & 63.22 & -- & 70.27 & 63.88 \\
\midrule
C $-$ B1 & +0.86 & $-$0.29 & +1.30 & -- \\
\;95\% interval & [$-$2.6, 3.6] & [$-$2.8, 1.5] & [$-$2.0, 4.9] & \\
C $-$ B0 & +1.15 & +0.96 & $-$0.14 & $-$3.08 \\
\;95\% interval & [0.0, 3.0] & [$-$0.6, 2.9] & [$-$3.4, 3.2] & [$-$6.0, 0.0] \\
\bottomrule
\end{tabular}}
\caption{The four pre-registered tests of the $t=3.0$ cascade (EX in \%; differences in points with database sign-flip intervals). RES: RESERVED; pool: three complete runs of every arm, EX averaged; External: KaggleDBQA test and Spider-Realistic; Fresh: new questions on post-cutoff databases (no B1 arm). Cost savings are in Figure~\ref{fig:forest}.}
\label{tab:tests}
\end{table}

\input{figures/fig_forest}
Routing queries between System~1 and the Claude Opus~5.5 agent via our pre-registered validator margin threshold ($t = 3.0$) answers roughly half of all questions locally and reduces API costs by 16--34\% with no significant loss of accuracy (Figure~\ref{fig:forest}, Table~\ref{tab:tests}). Across three complete runs on RESERVED (348 held-out BIRD questions), the cascade handled 49.1\% of queries locally on System~1 and cut total list-price cost by 34.3\% pooled against Opus with candidates (and 20.9\% against Opus alone in Run~1, including rented H100 GPU cost), while matching the agent's accuracy within statistical noise ($-0.29$ points pooled against Opus with candidates, $+0.96$ points against Opus alone). On our second pre-registered benchmark combining KaggleDBQA test and Spider-Realistic (693 questions across 27 databases), the cascade routed 51.9\% of questions to the agent, reducing cost by 31.7\% against Opus with candidates (+1.30 points accuracy) and 16.2\% against Opus alone ($-0.14$ points accuracy). While the first RESERVED run failed our pre-registered rule, missing its cost, time-saving and lower-bound bars (29.9\% vs.\ 30\%, 12.3\% vs.\ 25\%, and $-2.60$ vs.\ $-2.50$ points), the pooled runs and external benchmarks confirm that validator-margin routing consistently outperforms vote-agreement, question-length, and random routing baselines at the same call budget (Appendix~\ref{app:tests}).

\paragraph{Local hardware and direct API deployment.} When System~1 runs on hardware an organization already owns, its marginal cost is purely electricity; counting local compute as zero increases the cascade's cost savings against standalone Opus to 30.1\% pooled on RESERVED and 19.9\% on the external benchmark (Figure~\ref{fig:forest}). Furthermore, quantizing the shared backbone to 8-bit weights allows both adapters to run in about 9\,GB of unified memory on a 24\,GB Apple M3 laptop, matching GPU accuracy on 239 RESERVED questions (159 vs.\ 158 correct). Finally, when calling Opus directly through the plain Messages API, the $t = 3.0$ cascade reduces API cost by 33.3\% on our post-cutoff fresh benchmark at statistically indistinguishable accuracy (95.15\% vs.\ 95.59\%, $p = 1.0$); on RESERVED, measured afterwards, it saved 35.3\% but at 2.0 points lower accuracy ($p = 0.03$; Appendix~\ref{app:plainapi}).

\subsection{Benchmark Memorization vs.\ True Agent Generalization}
\label{sec:contam}
Because frontier LLMs may have memorized public text-to-SQL benchmarks during pretraining, we ran pre-registered contamination probes before trusting our System~2 baseline. Given only a question and database name---with no schema and no database tool---Opus~5.5 generated executable queries that returned the gold result on 79--95\% of instances across four public splits and reproduced 48--92\% of gold SQL queries verbatim, down to dataset-specific alias conventions and backtick formatting (Table~\ref{tab:probe} in Appendix~\ref{app:moreresults}). Standard schema-prompted evaluations on these public sets therefore mix reasoning with verbatim recall.

However, two controlled robustness tests suggest that Opus's high accuracy as a tool-using agent comes mostly from database exploration rather than memorization. First, when we systematically renamed every table and column in the RESERVED databases using a non-Claude model and paraphrased the questions, the routed agent's accuracy remained unchanged (72.88\% vs.\ 72.32\% on original schemas, $p = 1.0$). Second, on our newly constructed benchmark of 227 verified questions over 8 databases published after the agent's training cutoff (July--October 2026), System~1 answered 96.1\% of its kept questions correctly, the routed agent scored 92.80\% against 95.2\% for Opus alone on the same questions (the registered endpoint), and the full cascade achieved 94.27\% accuracy against 97.36\% for Opus alone (a 3.08-point gap, not significant) while cutting cost by 20.1\% (Table~\ref{tab:tests}).

%% file: figures/fig_forest.tex
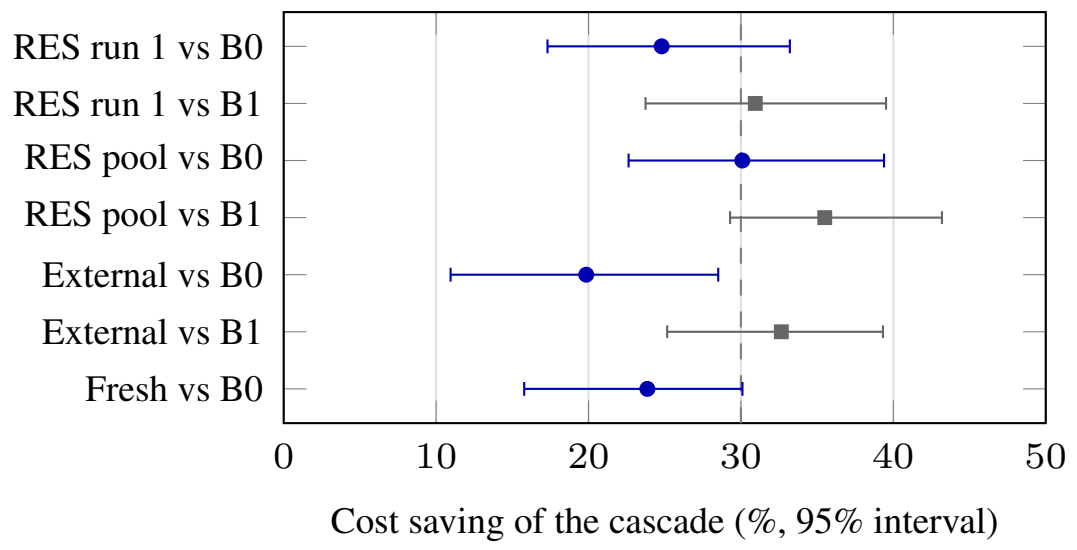
\begin{figure}[t]
\centering
\begin{tikzpicture}
\begin{axis}[width=0.86\columnwidth, height=4.3cm, xmin=0, xmax=50, ymin=0.4, ymax=7.6,
  xlabel={Cost saving of the cascade (\%, 95\% interval)}, ytick={1,2,3,4,5,6,7},
  yticklabels={Fresh vs B0, External vs B1, External vs B0, RES pool vs B1, RES pool vs B0, RES run 1 vs B1, RES run 1 vs B0},
  y tick label style={font=\scriptsize}, x tick label style={font=\scriptsize}, label style={font=\scriptsize},
  xmajorgrids, grid style={black!12}]
\draw[black!50, dashed] (axis cs:30,0.4) -- (axis cs:30,7.6);
\addplot[only marks, mark=*, mark size=1.3pt, blue!70!black, error bars/.cd, x dir=both, x explicit]
  coordinates {(24.80,7) -= (7.49,0) += (8.41,0)
               (30.09,5) -= (7.46,0) += (9.30,0)
               (19.85,3) -= (8.90,0) += (8.66,0)
               (23.86,1) -= (8.08,0) += (6.24,0)};
\addplot[only marks, mark=square*, mark size=1.2pt, black!60, error bars/.cd, x dir=both, x explicit]
  coordinates {(30.94,6) -= (7.20,0) += (8.58,0)
               (35.50,4) -= (6.21,0) += (7.69,0)
               (32.65,2) -= (7.49,0) += (6.67,0)};
\end{axis}
\end{tikzpicture}
\caption{Cost saving of the $t=3.0$ cascade in the held-out tests against Opus alone (B0, circles) and Opus with System~1's candidates on every question (B1, squares), with System~1 on local hardware. RES: RESERVED (348 BIRD questions; pool: three runs); External: KaggleDBQA and Spider-Realistic (693); Fresh: 227 post-cutoff questions. Dashed line: 30\%.}
\label{fig:forest}
\end{figure}

%% file: sections/conclusion.tex
\section{Conclusion}
\label{sec:conclusion}

Organizations deploying natural-language database interfaces do not have to choose between the accuracy of a frontier LLM agent and the speed, privacy, and zero marginal cost of a small local model. LuthorSQL, a one-pass slot-filling model, reaches 68.03\% execution accuracy on BIRD's hidden test set in 0.84\,s on a single GPU, and transfers zero-shot across six SQL engines without retraining. Placed in front of Claude Opus~5.5 as a fast System~1 gate, its validator margin routes roughly half of incoming queries locally and cuts end-to-end API costs by 16--34\% across pre-registered held-out benchmarks with no significant loss of the agent's accuracy. Finally, our contamination probes show that while frontier LLMs have memorized large portions of public text-to-SQL benchmarks verbatim, tool-assisted agents maintain high accuracy on obfuscated schemas and post-cutoff databases.

%% file: sections/limitations.tex
\section*{Limitations and Future Work}

\paragraph{Limitations.}
The grammar expresses 84.6\% of BIRD dev gold queries; questions needing CTEs, correlated subqueries, window functions or unseen templates usually go wrong, which is one reason to route them to an agent.
The cascade has not yet been run on BIRD's hidden test set, and its held-out tests are small (8, 27 and 8 databases), so intervals are wide and results vary between runs; RESERVED's first registered run failed its registered bars (Section~\ref{sec:s2results}). Time savings hold against Opus with candidates but not against Opus alone, since the routed questions are the slow ones, and agent time depends on load.
Costs are list prices applied to logged usage, not invoices, and the agents ran inside Claude Code, whose fixed per-session cost makes part of the saving come from running fewer sessions; through the plain API the saving is registered on the fresh set only.
Opus has probably seen every public set we used (Section~\ref{sec:contam}); the renamed-schema test blocks verbatim recall but not recall of a query's structure, and the fresh set is easier than BIRD and was checked by LLM agents and spot-checked by hand rather than reviewed in full.
On BIRD dev the agent is much stronger on confident questions, so the $t=3.0$ threshold lost 2.93 points there; on DECISION, out-of-grammar questions were routed by gold coverage, which the held-out tests avoid. These questions were identified by converting their gold SQL, information a deployed system would not have, so the DECISION cascade results assume a perfect out-of-grammar detector, and in the handoff the agent saw an empty list on these questions (70 of the 400) instead of System~1's usually wrong candidates, which can mislead it; this may flatter the handoff's accuracy and change its time savings. DECISION and RESERVED's databases are in OmniSQL's fine-tuning data. In designing \textsc{VALHARD}, the hard-negative perturbation categories were chosen after inspecting an earlier validator's errors on DECISION, the same split used to evaluate \textsc{VALHARD}, so its +1.76-point gain on DECISION may be optimistic (the BIRD dev score of 59.39\% played no part in its design). Furthermore, \textsc{VALHARD} narrowly missed our pre-written database-level significance bar on DECISION (database sign-flip $p = 0.19$, question-level McNemar $p = 0.04$), though we retained it as no subsequent validator modification outperformed it (Appendix~\ref{sec:negative}).
BIRD's test score is 8.64 points above our dev score and we cannot say why without test labels; it comes from BIRD's emailed result and is not yet on the public leaderboard.
Everything here is English with foreign-key joins; System~1 runs on MySQL, PostgreSQL, DuckDB, Snowflake and BigQuery through translation of its SQLite output rather than natively, the cascade's agent was tested on SQLite only, and the system needs read access to the database.

\paragraph{Future work.}
Next steps are the $t=3.0$ cascade on BIRD's hidden test set, a plain-API cost measurement, native generation in other SQL dialects, caching the validator's schema pass per database (about half of System~1's time), and a detector that routes out-of-grammar questions before System~1 runs.

\section*{Ethics Statement}
All training and evaluation data are public; the evaluation sets are KaggleDBQA (CC BY-SA 4.0), Spider-Realistic (built from Spider dev, CC BY-SA 4.0) and the fresh set, whose sources' licenses are in Appendix~\ref{app:fresh}. The reported system uses BIRD (CC BY-SA 4.0), Spider (CC BY-SA 4.0) and SynSQL-2.5M (Apache-2.0); earlier rounds also used WikiSQL \citep{zhong2017seq2sql} (BSD-3-Clause), Spider-Syn \citep{gan2021spidersyn} (MIT), SParC and CoSQL (both CC BY-SA 4.0). OmniSQL-7B and Qwen2.5-14B-Instruct \citep{qwen2024qwen25} (the model that wrote R1's rare-skeleton questions) are released under Apache-2.0.
The BIRD-Platinum labels of \citet{revisql2026}, for which we found no license, were used for research only: in VALHARD's training lists, which makes them part of the reported system, and in other validator and judge training and one generator experiment.
No private or proprietary data was used.
LLM outputs (answers from Opus~5.5, Sonnet, Haiku, GLM~5.3, GLM~5.3 Flash and DeepSeek V4.1 Flash, and judge verdicts) were used for evaluation, and some training questions were written by LLMs: DeepSeek V4.1 Flash and Claude subagents in rounds 2--3, DeepSeek V4.1 Flash for one negative-result experiment, and Qwen2.5-14B for R1's rare skeletons.
A text-to-SQL system can return a confident wrong answer to a user who cannot read SQL; the validator margin used by the cascade is one signal for flagging such cases, not a guarantee.
Ibrahim Memon designed and ran the experiments and wrote the paper; Armin Karamzade gave feedback on the framing and presentation of the paper; Hanif Jhaveri advised on the work; Roy Fox supervised the work. An AI assistant (Claude, through Claude Code) wrote much of the experiment code under the first author's direction and helped prepare the manuscript from the first author's notes and experiment logs; the first author revised the text and checked every number against those logs, and all authors take responsibility for the content.

%% file: sections/history.tex
\section{How We Got Here}
\label{sec:history}

\begin{table*}[t]
\centering\small
\setlength{\tabcolsep}{4pt}
\begin{tabular}{@{}llp{4.1cm}cp{5.6cm}@{}}
\toprule
Date (2026) & Stage & Model & BIRD dev & What it taught us \\
\midrule
Sep 30 & Round 1 & 4B bi-encoder fills a plan & 19.3 & Wrong output column caused 48--61\% of errors; question and schema never met inside the model. \\
Oct 1--3 & Rounds 2--3 & 4B, schema in context, reranker & -- & Grammar v2 covers 70.9\% of dev; best held-out system 73.3\% against a top-8 ceiling of 77.3\%. \\
Oct 3--4 & Rounds 4--5 & 7B autoregressive SQL writer & 63.2$^\ast$ & Writing SQL beat filling slots on the same held-out list (first answer 83.4 against 78.4 for the slot model); a trained selector lost 2.7 points to plain voting. \\
Oct 4 & Round 6 & OmniSQL-32B, three prompts & 68.3$^\dagger$ & A 32B track would at best edge past open 32B models; we went back to one pass. \\
Oct 4 & Slot pilot & OmniSQL-7B slot model & 37.5 & Grammar v3 lifts coverage to 84.6\%; 42\,ms per question. \\
Oct 5--7 & R1, R2 & + validator, evidence repair & 56.78/56.71 & Validator over top 8 and a deterministic repair; doubling the training steps changed nothing. \\
Oct 8 & ENS1 & two runs pooled & 59.06 & The gain came from a larger candidate pool, not from voting validators. \\
Oct 9--11 & System~1 & R1, $k{=}32$, VALHARD & \textbf{59.39} & One generator and one validator score about the same as the pooled pair; frozen and scored once. \\
Oct 12 & System~1+2 & + Opus~5.5 agent & -- & Handoff and cascade measured on held-out training databases (Appendix~\ref{app:selection}). \\
Oct 18 & System~1+2 & all of dev, threshold on 796 & 84.49$^\ddagger$ & Agent with candidates; the cascade threshold from 400 questions failed on dev and was re-chosen on all 796 ($t=3.0$). \\
\bottomrule
\end{tabular}
\caption{Milestones, with BIRD dev execution accuracy (\%, original gold, all 1,534 questions unless marked). $^\ast$Execution vote over greedy plus 16 samples; greedy 57.8. $^\dagger$Corrected 2025 gold, vote over three prompt styles; not comparable to the other rows. Rounds 2--4 were not scored on BIRD dev; the Rounds 4--5 row gives round 5's score. $^\ddagger$With Opus~5.5 as the agent on every question; possibly partly inflated by memorization (Section~\ref{sec:contam}).}
\label{tab:history}
\end{table*}

All stages used public data only. The first version, a bi-encoder on the \texttt{openjev} Qwen3.5-4B model \citep{wortega2026openjev}, encoded question and columns separately; the wrong output column caused 48--61\% of its errors, so later versions put the schema in front of the question. An autoregressive detour (Qwen2.5-Coder-7B, \citealp{hui2024qwen25coder}; OmniSQL-32B) scored higher but cost seconds per question, so we returned to one pass on OmniSQL-7B with grammar v3, which covers 1,297 of 1,534 dev gold queries (84.6\%). Pooling two runs' candidates gave 59.06\%; the gain came from the larger candidate list, so we kept one run (R1) with up to 32 distinct plans and retrained its validator on hard negatives (VALHARD). Because no further selection change passed its bar (Appendix~\ref{sec:negative}) and 15\% of dev questions lie outside the grammar, where a model that writes SQL freely does better (35.9\% against 14.8\% for our system on those 237 questions), we paired System~1 with an agent instead of growing the grammar.

%% file: sections/negative.tex
\section{What Did Not Work}
\label{sec:negative}

\begin{table}[t]
\centering\footnotesize
\setlength{\tabcolsep}{1.5pt}
\begin{tabular}{@{}lrl@{}}
\toprule
Attempt & $\Delta$ EX & Verdict \\
\midrule
\multicolumn{3}{@{}l}{\emph{Selection}} \\
LLM judge, top 4 (think off/on) & $-$0.25/+0.63 & no bar \\
8 decision-model APIs & $-$2.39 to +1.88 & no pass \\
Two-judge agreement & +2.01 & RES.\ +0.29 \\
Fine-tuned 27B judge & +0.75 & no pass \\
Gated pairwise duel & +0.50 & kill \\
Third validator in the pool & $-$0.38 & kill \\
Retrained validator (VAL2) & $-$0.38 & kill \\
Results-aware validator & $-$0.75 & kill \\
VAL3 recipe / joint scoring & $-$1.01/$-$4.15 & kill \\
VAL4, unseen-database lists & $-$0.13 & kill \\
Free pick rules, same pool & $-$0.57$^\dagger$ & no pass \\
\midrule
\multicolumn{3}{@{}l}{\emph{Generation}} \\
RL, GRPO/pass@8 & +0.13/$-$0.25 & kill \\
LoRA soup of two runs & $-$1.26 & kill \\
Decode-time constraints & $-$0.50 to +0.13 & drop \\
+33k LLM-written rows & +0.50 & kill \\
Corrected training labels & $-$0.75 & kill \\
Grammar v4 + descriptions & $-$1.26 & kill \\
Doubling training steps & $-$0.07$^\ddagger$ & flat \\
Question-variant ensemble & +0.63 & no pass \\
\midrule
\multicolumn{3}{@{}l}{\emph{One adapter for both jobs}} \\
Soup of runs 1 and 2 & $-$1.76 & miss \\
2-layer rerank head & $-$1.01 & miss \\
\bottomrule
\end{tabular}
\caption{Attempts measured on DECISION (796 questions, original gold) against the system current at the time, with bars written before scoring except for the LLM judge (\emph{no bar}: exploratory, no gain) and seven of the eight decision-model APIs (exploratory). RES.\ = the same change scored once on RESERVED. $^\dagger$On RESERVED. $^\ddagger$On BIRD dev (original gold). The single-adapter rows are against the two-adapter LuthorSQL (Table~\ref{tab:single}). }
\label{tab:negative}
\end{table}

Table~\ref{tab:negative} lists the attempts that did not survive; their scripts, costs and raw outputs are in the code release. No selection change after VALHARD passed: off-the-shelf and fine-tuned judges did not beat the trained 7B validator by the per-database test, and a judge combination that gained 2.01 points on DECISION fell to +0.29 on RESERVED after we had tried about eleven configurations on the same questions, so the first number mostly measured our search. No generation change passed either: doubling the training steps, execution-reward RL, extra LLM-written rows, corrected labels and a larger grammar all landed within noise or below. Two seeds of one recipe differ by 0.50 points on DECISION, which is why single-run generator claims needed +2.0.

%% file: sections/app_single.tex
\section{One adapter for generation and selection}
\label{sec:single}

BIRD judged that two separately trained adapters belong on its Overall leaderboard and invited a single-adapter resubmission for the Single Trained Model track, so we built one. It keeps R1's rank-32 adapter and slot heads and adds a zero-initialized linear rerank head on the same adapter's last hidden state, so no second adapter exists. Each training step combines one VALHARD validator batch with one generator chunk replayed from R1's training mix; no DECISION, RESERVED or dev question is in the training data, and inference is unchanged with one backbone in memory.

\begin{table}[t]
\centering\footnotesize
\setlength{\tabcolsep}{1.5pt}
\begin{tabular}{@{}lcccccc@{}}
\toprule
System & EX & $\Delta$ & DB $p$ & McNemar & Or.@32 & Sel. \\
\midrule
Two adapters & 59.05 & -- & -- & -- & 67.84 & 59.05 \\
Run 1 & 58.29 & $-$0.75 & 0.594 & +22/$-$28 & 67.59 & 58.42 \\
Run 2 & \textbf{58.54} & $-$0.50 & 0.750 & +24/$-$28 & \textbf{68.59} & 58.42 \\
Soup & 57.29 & $-$1.76 & 0.156 & +15/$-$29 & 67.59 & 57.91 \\
2-layer head & 58.04 & $-$1.01 & 0.445 & +20/$-$28 & 67.59 & \textbf{58.67} \\
\bottomrule
\end{tabular}
\caption{Single-adapter arms on DECISION (796 questions, original gold; out-of-grammar questions count as wrong): runs 1 and 2, their weight average (soup) and run 1's recipe with a 2-layer rerank head. $\Delta$ is against the two-adapter LuthorSQL, with the database sign-flip $p$ and McNemar wins/losses. Or.@32: oracle over the arm's own list of up to 32 candidates. Sel.: the arm's scorer picking from R1's own candidates.}
\label{tab:single}
\end{table}

Run 1 (2 epochs, generator weight 1.0) and run 2 (3 epochs, weight 0.5, chosen by a rule written before run 1 was scored) are 0.75 and 0.50 points below the two-adapter system, not significantly (Table~\ref{tab:single}), though the intervals are wide (run 2: $[-4.06, +2.74]$), so this is not evidence of equal accuracy. The gap is mostly in selection: on R1's candidates both pick at 58.42\% against VALHARD's 59.05\%. A weight average (soup) of the two runs and a 2-layer rerank head did worse. We resubmitted run 2 to BIRD for its Single Trained Model track. On the hidden test set it scores 68.19\% EX (simple 77.03, moderate 65.77, challenging 43.51), 69.35\% Soft F1 and 64.58\% R-VES, against 68.03\%, 69.26\% and 65.17\% for the two adapters, so one adapter loses no test accuracy. On BIRD dev (original gold, scored only after the test result came back) it scores 59.84\% (918/1,534) against 59.39\%.

%% file: sections/app_valhard.tex
\section{VALHARD training data}
\label{app:valhard}
We built VALHARD after reading the earlier validator's misses on held-out training databases.
Of the 84 questions where it missed although a right candidate was in its list, the right result was its second choice for 58; the clear error kinds were counting at the wrong grain, a dropped or wrong filter, join, literal or sort, and the wrong output columns (34 of the 84 by an automatic bucketing; 11 more were LIMIT or tie cases and 39 were unclassified).
Its training lists come from the generator's own $k{=}32$ candidates on 42 BIRD-train databases (3,830 lists, with the expert-corrected BIRD-train labels of \citet{revisql2026} (BIRD-Platinum) where available), with one live representative per result signature, and two sources of hard negatives: the closest right/wrong pair for each wrong result class (28,450 mined, 16,804 of them in the training lists), and synthetic sqlglot edits of right queries of the same kinds (count versus count distinct, an added join, a dropped filter, a flipped order, a dropped or added projection), each kept only if it runs and returns a new, wrong, non-null result.
The loss treats each result group as one option, scoring it by the log-mean-exp of its members.

%% file: sections/app_setup.tex
\section{Training and measurement details}
\label{app:setup}
\paragraph{Training.}
All training data is public: BIRD train \citep{li2023bird}, Spider train \citep{yu2018spider} and a sample of SynSQL-2.5M \citep{li2025omnisql}; each gold query is kept only if its compiled plan returns the gold result. We add ``twin'' questions that differ from an original in one word and have a different answer, and 5,483 questions written by Qwen2.5-14B-Instruct \citep{qwen2024qwen25} for rare plan skeletons; BIRD train is repeated three times.
R1 uses AdamW (learning rate $10^{-4}$ for the LoRA weights, rank 32, $\alpha{=}64$, all linear layers; $10^{-3}$ for the heads), cosine decay and bf16, with 8 questions from each of 4 databases per step. It ran 6,714 steps in 6.7\,h on one A100; we keep the step-6,000 checkpoint, the best by greedy accuracy on 1,455 held-out BIRD-train questions. VALHARD trained in about 2.1\,h on one H100.

\paragraph{Latency.}
200 seeded DECISION questions at batch size 1 after three warm-up questions, counting every stage and excluding model load and one-off per-database setup, on one H100 PCIe and on an A100 80\,GB with a CUDA 12.2 driver (BIRD's environment). Dollar figures use \$2.50/h (H100) and \$2.00/h (A100). On a 24\,GB Apple M3 laptop (PyTorch on the Apple GPU), both adapters share one backbone whose linear layers hold 8-bit weights with one scale per output channel; LoRA weights, embeddings and heads stay bf16. The run covered 239 RESERVED questions on 6 of the 8 databases, with other applications open, before memory pressure from those applications stopped it; it gave the GPU run's answer on 207 of them (87\%), 159 right against 158. In bf16 the same set-up peaks at 17.2\,GB.

\paragraph{System~2 protocol.}
\label{sec:setup2}
For the Claude agents the eight-query limit was an instruction: they sometimes batched several queries into one tool call, so compliance could not be checked exactly. The open-weight API agents in Appendix~\ref{app:selection} had the limit enforced in code. Each chunk of up to 25 questions is one Claude Code subagent with the SQL tool and the prompts of Appendix~\ref{app:prompts}; tokens and time come from the subagents' usage logs. Costs are Opus~5.5 list prices (input \$4/M tokens, five-minute cache writes \$5/M, cache reads \$0.20/M, output \$20/M, output taken as the larger of logged tokens and written characters / 4), not invoices. The DECISION handoff comparison used a fixed 400-question sample with the two arms interleaved; the threshold was then chosen on all 796 after a second handoff run on the other 396. The Opus agent-alone DECISION score (77.26\%) comes from an earlier October 10 run; in the Opus DECISION runs gold SQL sat in the agent's repository behind an instruction only, while all later runs (dev, other agents, held-out tests) used a sandbox holding only the databases and the tool, and their transcripts were scanned for access outside it. GLM~5.3, GLM~5.3 Flash and DeepSeek V4.1 Flash ran through the Ollama Cloud API in a plain tool loop at temperature 0, one question per conversation, with the eight-execution limit enforced in code.

\paragraph{Held-out tests.}
RESERVED's registration fixed the endpoints, the bars (cost saving $\ge 30\%$ against Opus with candidates with a bootstrap lower bound $\ge 20\%$; accuracy lower bound $\ge -2.5$ points; time saving $\ge 25\%$) and predictions before any agent ran. System~1 ran gold-free on every question on a rented H100 (\$0.00085 per question including model load), and the routing share was recorded before the cascade arm ran. Each arm ran in chunks from a single database, in interleaved waves; intervals resample databases 10,000 times.

\paragraph{Splits and dev looks.}
DECISION and RESERVED come from 8 held-out BIRD-train databases (hockey, student\_loan, synthea, university, car\_retails, law\_episode, cs\_semester, movie), split by database, grammar coverage and gold problems, 30\% of each group to RESERVED. RESERVED was scored by System~1 during development (60.34\% with out-of-grammar questions counted as wrong; 63.22\% in Table~\ref{tab:tests}, where the gold-free package answers them too), so it is not a clean final test for System~1, though never with an agent. R1's checkpoint was selected on these databases, and many of their questions were in an earlier held-out set used before October 8 (Appendix~\ref{app:leak}). From October 8 we allowed at most three BIRD dev looks: LuthorSQL used the first, the System~2 dev run the second.

\paragraph{Scoring.}
System~1 scores (including the dev look) use a 3\,s limit per query and at most 100,000 rows; agent experiments use 30\,s (BIRD dev: 60\,s) and no row cap. Bars were written before scoring for the registered tests only; the DECISION handoff and cascade runs were exploratory, and the threshold was chosen by a rule written before the full run.

%% file: sections/app_results.tex
\section{Additional results}
\label{app:moreresults}
\label{app:selection}

\begin{table}[t]
\centering\small
\setlength{\tabcolsep}{2.5pt}
\begin{tabular}{lcccc}
\toprule
Mode (R1 generator) & DEC. & Dev & H100 & A100 \\
\midrule
First plan only & 49.12 & -- & 0.13\,s & 0.18\,s \\
$k{=}32$ + VALHARD & 59.05 & \textbf{59.39} & \textbf{0.84\,s} & 1.23\,s \\
$k{=}64$ + repair + val. & 59.55 & -- & 1.60\,s & 2.37\,s \\
\midrule
\multicolumn{5}{l}{Batched, $k{=}32$ + VALHARD} \\
\;questions / s & & & 1.09 & 0.77 \\
\;\$ per 1,000 questions & & & 0.64 & 0.72 \\
\bottomrule
\end{tabular}
\caption{Speed tiers of the two-adapter System~1 pipeline (bold: the frozen LuthorSQL configuration). EX (\%) on DECISION (796 held-out training questions) and BIRD dev (original gold, all 1,534); times are p50 per question at batch size 1, end to end, measured on 200 DECISION questions with model load excluded (H100 PCIe and A100 80GB PCIe); cost at \$2.50/h (H100) and \$2.00/h (A100).}
\label{tab:latency}
\end{table}

\paragraph{System~1 alone.}
LuthorSQL scores 59.39\% on BIRD dev (911 of 1,534; three gold queries fail to run and count as wrong), with 69.41\% on simple, 45.69\% on moderate and 39.31\% on challenging questions, and 55.61\% on the 2025 corrected gold, though it saw the original question text.
One generator and one validator score about the same as the earlier system that pooled two runs (59.06\%), and against OmniSQL-7B (62.58\%) that pooled system was level on the 85\% of dev questions the grammar covers and 21 points behind on the rest (14.77\% against 35.86\%). On an A100 the package picks the same SQL as the frozen run on 170 of 183 questions, with accuracy equal to within one question.

\begin{table}[t]
\centering\small
\setlength{\tabcolsep}{3pt}
\begin{tabular}{lccc}
\toprule
Model & Greedy & Vote/pick & Latency (s) \\
\midrule
LLaDA2.1-mini (DLM) & 48.8 & 58.7 & 5.66 \\
WeDLM-8B (DLM) & 51.4 & 53.0 & 0.18 \\
Qwen2.5-Coder-7B (AR) & 50.9 & 58.2 & 0.31 \\
Qwen2.5-Coder-14B (AR) & 61.5 & 66.1 & -- \\
\midrule
OmniSQL-7B (AR)$^\ast$ & 62.6 & -- & -- \\
Our AR writer (7B)$^\ast$ & 57.8 & 63.2$^\S$ & -- \\
LuthorSQL$^\ast$ & -- & 59.39$^\|$ & 0.84 \\
\bottomrule
\end{tabular}
\caption{BIRD dev EX (\%, original gold). The first block is copied from \citet{ma2026sqld1}, an unreviewed preprint that does not say whether BIRD's evidence was given (v1, Tables 1 and 7, unchanged in v2's Tables 2 and 10; latency on A100s as they report it), prompted with retrieval-augmented schema context in a setting they call zero-shot. $^\ast$Our runs, given BIRD's evidence; the writer and LuthorSQL are fine-tuned on public training data, and OmniSQL-7B is the released checkpoint decoded greedily. Vote/pick: majority vote over 8 samples (Maj@8) unless marked. $^\S$Vote over 17 candidates. $^\|$Validator pick from up to 32 distinct candidates, not a vote; latency is the p50 on one H100 PCIe over 200 DECISION questions, not timed on BIRD dev.}
\label{tab:dlm}
\end{table}

Table~\ref{tab:dlm} places LuthorSQL next to the diffusion and autoregressive models evaluated by \citet{ma2026sqld1}: it is roughly level with their best diffusion result (58.7\%, an 8-sample vote; they report a latency of 5.66\,s without saying whether per question, against our 0.84\,s per question), under different conditions and hardware, while stronger SQL-writing models remain ahead: Qwen2.5-Coder-14B at 66.1 with an 8-sample vote (61.5 as a single answer) in their runs, and OmniSQL-7B at 62.6 in ours. On BIRD dev the package's pick differed from the scored run on 59 of 1,534 questions (3.8\%), which bounds the implementation drift there.

\begin{table}[h]
\centering\small
\setlength{\tabcolsep}{2.5pt}
\resizebox{\columnwidth}{!}{\begin{tabular}{@{}lcccc@{}}
\toprule
 & Spider-R. & Kaggle & BIRD tr. & Mini-Dev \\
\midrule
Questions & 508 & 185 & 348 & 500 \\
EX, no schema & 94.69 & 79.46 & 79.31 & 86.20 \\
Gold SQL verbatim & 92.3\% & 53.0\% & 48.3\% & 58.4\% \\
EX with database (B0) & 92.72 & 76.22 & 80.17 & -- \\
\midrule
Exact completions & 22 / 16 & 66 / 4 & 77 / 8 & 91 / 25 \\
\bottomrule
\end{tabular}}
\caption{Memorization probes on Opus~5.5 (no database, no tool). EX, no schema: given only the question and the database name; verbatim: the gold query reproduced after whitespace and quote normalization. Exact completions: of 100 half-questions, second halves reproduced word for word when told / not told the dataset name. BIRD tr.: RESERVED; Mini-Dev: the SQLite version of BIRD Mini-Dev, drawn from BIRD dev, scored on BIRD dev's databases rather than Mini-Dev's own copy (2 of its 500 gold queries failed to run there and count as wrong).}
\label{tab:probe}
\end{table}

\paragraph{Other SQL dialects.}
\label{app:dialects}
System~1 compiles plans to SQLite. To run it on other engines we translate every candidate with sqlglot (SQLite to the target dialect), match table and column names to the target catalog without regard to case, qualify tables with their schema or dataset on Snowflake and BigQuery, and, for PostgreSQL, add type rules for what SQLite coerces implicitly (string literals against text columns, dates read as text for \texttt{LIKE}, \texttt{SUBSTRING} and comparisons, \texttt{LIKE} as \texttt{ILIKE}, \texttt{JULIANDAY} through the epoch); these rules were written after reading PostgreSQL errors on Mini-Dev. Translated candidates are executed on the target database and picked by the package's rule (the validator reads SQL text, so its scores carry over). No model is retrained or rerun: we reuse the frozen package run on dev. BIRD's MySQL and PostgreSQL dumps hold all 11 dev databases, built from BIRD's updated dev files (5 of 11 differ from the 2024-06-27 release), so every engine is scored on those files: we copy the same SQLite files into native DuckDB files, into one Snowflake schema and one BigQuery dataset per database (columns SQLite stores with mixed types become text; 11 BigQuery column names with characters it rejects are renamed and mapped back); on all 1,534 questions a prediction is right when its rows equal the SQLite gold rows after normalizing values (numbers rounded to 6 places, dates as ISO strings). On Mini-Dev, where BIRD publishes gold for each dialect, the official rule gives 50.4\% (SQLite), 49.2\% (MySQL) and 48.8\% (PostgreSQL; 48.4\% without the type rules).

\begin{table}[h]
\centering\small
\setlength{\tabcolsep}{3pt}
\begin{tabular}{@{}lcccc@{}}
\toprule
BIRD dev (1,534) & EX & Simple & Mod. & Chall. \\
\midrule
SQLite & 60.17 & 70.16 & 46.77 & 39.31 \\
Snowflake & 59.32 & 69.30 & 46.12 & 37.93 \\
MySQL & 57.63 & 67.14 & 44.83 & 37.93 \\
DuckDB & 57.43 & 67.68 & 43.97 & 35.17 \\
PostgreSQL & 57.37 & 66.59 & 45.47 & 36.55 \\
BigQuery & 57.11 & 67.14 & 43.10 & 37.93 \\
\bottomrule
\end{tabular}
\caption{System~1 on other engines without retraining. Share of candidates that run: MySQL 96\%, Snowflake 83\%, PostgreSQL 83\%, DuckDB 80\%, BigQuery 76\%; stricter GROUP BY rules and typing account for most failures.}
\label{tab:dialects}
\end{table}

\begin{table}[t]
\centering\small
\setlength{\tabcolsep}{3pt}
\begin{tabular}{lccc}
\toprule
 & Opus alone & + System~1 & Change \\
\midrule
EX (400 questions) & 77.75 & 78.00 & +0.25 \\
Agent time, total & 1,587\,s & 1,146\,s & $-$28\% \\
Time, median chunk & 91\,s & 56\,s & $-$38\% \\
Tool calls (all) & 207 & 169 & $-$18\% \\
Output text (chars/4) & 54.7k & 33.3k & $-$39\% \\
Cache writes & 842k & 1,025k & +22\% \\
Cache reads & 11.8M & 10.8M & $-$9\% \\
\$/question (list) & $\approx$0.019 & $\approx$0.020 & $\approx$+4\% \\
\bottomrule
\end{tabular}
\caption{Candidate handoff on a 400-question DECISION sample. Times and tokens include the same agent-harness overhead in both arms. Output text is an estimate (output characters / 4); logged output-token counts were not lower.}
\label{tab:handoff}
\end{table}

\paragraph{Handoff (DECISION, exploratory).}
Given System~1's candidates, the agent shows no measurable change in accuracy (78.00\% against 77.75\%; $p=1.0$) and does less work: 28\% less agent time and 18\% fewer logged tokens. Of the 330 questions with candidates it took one on 279 (87\% right) and wrote its own query on the other 121 (57\% right); in an earlier pilot, on questions with no right candidate it scored 43.2\% with candidates against 47.2\% alone, so wrong candidates can mislead it. The longer prompt cancels the saving from shorter answers, so the cost per call stays at about 2 cents; the money is saved by not calling the agent.

\begin{table}[t]
\centering\small
\setlength{\tabcolsep}{4pt}
\begin{tabular}{cccc}
\toprule
Margin $\ge t$ & EX & Agent calls & Change \\
\midrule
always agent & 78.89 & 100\% & -- \\
2.0 & 77.76 & 45.6\% & $-$1.13 \\
2.5 & 78.14 & 49.5\% & $-$0.75 \\
\textbf{3.0} & \textbf{78.39} & \textbf{54.6\%} & $\mathbf{-0.50}$ \\
3.5 & 78.27 & 59.3\% & $-$0.63 \\
4.0 & 78.52 & 66.6\% & $-$0.38 \\
5.0 & 78.77 & 85.8\% & $-$0.13 \\
\bottomrule
\end{tabular}
\caption{Confidence cascade on all 796 DECISION questions. System~1 answers alone when its validator margin is at least $t$; otherwise the agent gets the handoff prompt. System~1 alone scores 58.92 here: the cascade picks among result groups, which breaks one tie differently from the official pick (59.05). Opus alone scores 77.26. Bold: the threshold picked by the rule fixed before the run, which charged every agent call the same average list-price cost, estimated at the time; measured held-out results are in Table~\ref{tab:tests} and Figure~\ref{fig:forest}.}
\label{tab:cascade}
\end{table}

\paragraph{Cascade threshold (DECISION).}
The rule picks $t=3.0$: System~1 answers 45.4\% of the questions alone and the combined system scores 78.39\%, 0.50 points below always calling the agent ($p=0.50$, $[-1.44, +0.50]$), with 54.6\% of the agent calls (Table~\ref{tab:cascade}). Here the 126 out-of-grammar questions were routed by gold coverage; routing by the margin alone, as in the held-out tests (402 of 796 routed), gives 77.64\%, 1.25 points below always calling the agent (both reuse the agent's earlier answers, whose candidate lists were empty for those questions). Because the threshold was picked on these questions, 78.39\% is an in-sample estimate; between 2.25 and 4.5 the curve stays within four questions, and leaving out one database at a time the rule still picks 3.0 on six of eight folds. The threshold of 2.5, picked earlier on the 400-question sample, failed on BIRD dev (below), so we re-chose it on all 796; RESERVED and the external sets played no part.

\begin{table}[t]
\centering\small
\setlength{\tabcolsep}{3pt}
\begin{tabular}{lcccc}
\toprule
Agent & Alone & + S1 & $t{=}3.0$ & Change \\
\midrule
Opus~5.5 & 77.26 & 78.89 & 78.39 & $-$0.50 \\
Sonnet & 71.11 & 74.50 & 74.37 & $-$0.13 \\
Haiku & 68.34 & 72.61 & 72.61 & 0.00 \\
\midrule
GLM~5.3$^\ast$ & 66.21 & 70.48 & 70.60 & +0.13 \\
GLM~5.3 Flash$^\ast$ & 65.95 & 68.47 & 68.84 & +0.38 \\
DeepSeek V4.1 Fl.$^\ast$ & 65.95 & 69.22 & 70.23 & +1.01 \\
\bottomrule
\end{tabular}
\caption{The same cascade with six agents (all 796 DECISION questions; System~1 alone 58.92). + S1: the agent with System~1's candidates on every question; $t{=}3.0$: the cascade; Change: cascade minus + S1. The agent sees the same 54.6\% of questions in every row, since routing does not depend on the agent; the $t{=}3.0$ column combines System~1's answers on the kept questions with the + S1 arm's answers on the routed ones, rather than being a separate run. $^\ast$Open-weight models run through Ollama Cloud with a plain API loop (one question per conversation, schema in the first message, the eight-query limit enforced in code) instead of Claude Code subagents, so their absolute scores are not directly comparable with the Claude rows. Fl.: Flash.}
\label{tab:agents}
\end{table}

\paragraph{Other agents (DECISION).}
To check that the threshold is not tied to Opus we repeated both arms with five agents weaker than Opus on this split: Claude Sonnet~5.5 and Haiku~5.5 (\texttt{claude-sonnet-5-5}, \texttt{claude-haiku-5-5}), GLM~5.3, GLM~5.3 Flash and DeepSeek V4.1 Flash (Table~\ref{tab:agents}). Each gains +2.51 to +4.27 points from the candidates (significant at the database level only for Haiku), and at $t=3.0$ the largest observed loss against always calling the agent is 0.13 points (only DeepSeek's +1.01 is significant). Above a margin of 3.5 the open models are up to 3.7 points below System~1, so sending them confident questions hurts. Our predictions, written before these runs and committed with the results, failed for the chosen threshold of Haiku (2.25) and the open models (2.0, on curves flat within 0.3 points from 2.0 to 4.0) and for GLM~5.3 Flash's predicted gain of at least 10 points (+9.55); GLM~5.3 was added after the registration. What held is that $t=3.0$ costs little with every agent we tried.

\begin{table}[t]
\centering\small
\setlength{\tabcolsep}{3pt}
\begin{tabular}{lccc}
\toprule
Arm & Agent calls & EX & Corrected \\
\midrule
System~1 alone & 0\% & 59.39 & 55.67$^\ast$ \\
Opus alone & 100\% & 84.03 & 77.25 \\
Opus + System~1 & 100\% & \textbf{84.49} & 76.27 \\
Cascade $t{=}2.5$ & 49.5\% & 80.70 & 72.88 \\
Cascade $t{=}3.0$ & 55.0\% & 81.55 & 73.66 \\
\bottomrule
\end{tabular}
\caption{All 1,534 BIRD dev questions (our second dev look). EX uses the original gold; Corrected rescores the same answers against BIRD's 2025 revision of the same question ids and is not a leaderboard number. $^\ast$55.61 under the 3\,s System~1 scorer (Section~\ref{sec:results}); this table uses the 60\,s scorer with no row cap; the two differ by one question, which we have not traced. $t{=}2.5$ was the threshold pre-registered for this run from the 400-question sample; $t{=}3.0$ is shown for reference and was chosen after.}
\label{tab:dev2}
\end{table}

\input{figures/fig_bands}

\paragraph{BIRD dev.}
On all 1,534 dev questions, Opus with System~1's candidates scores 84.49\%, 25.10 points above System~1 alone and not significantly different from Opus alone (84.03\%; $p=0.70$). The cascade saves less there: at $t=3.0$ it scores 81.55\%, and the registered $t=2.5$ scored 80.70\%, 3.78 points below always calling the agent, failing its non-inferiority bar. System~1's confidence transferred (its accuracy by margin band is close to DECISION's above 2.5), but the agent's gain did not: above 2.5 Opus adds 1.2--2.6 points per band on DECISION and 12--15 on dev (Figure~\ref{fig:bands}). Part of that may be recall: 20.9\% of Opus's answers alone match the gold SQL text, and on 83 dev questions whose gold result changed in BIRD's 2025 revision, Opus with candidates leans to the old gold (39 against 26) while System~1 leans to the new (24 against 15), though with 83 questions none of these splits is significant. The lesson for a cascade is to measure the agent's gain over the small model in each confidence band on questions the agent has not seen.

%% file: figures/fig_bands.tex
\begin{figure}[t]
\centering
\begin{tikzpicture}
\begin{axis}[width=\columnwidth, height=5.2cm, ymin=0, ymax=100, ylabel={EX (\%)}, xlabel={System~1 validator margin},
  xtick={0,...,8}, xticklabels={none,0--0.5,0.5--1.4,1.4--2.5,2.5--3.5,3.5--4.1,4.1--4.6,4.6--5.4,$\ge$5.4}, x tick label style={font=\tiny, rotate=40, anchor=north east},
  y tick label style={font=\scriptsize}, label style={font=\scriptsize}, ymajorgrids, grid style={black!12},
  legend style={font=\tiny, at={(0.98,0.03)}, anchor=south east, draw=none, fill=white, fill opacity=0.85, text opacity=1}, legend cell align=left]
\addplot[black, thick, mark=*, mark size=1.3pt] coordinates {(0,4.6) (1,33.8) (2,43.4) (3,58.4) (4,79.0) (5,84.1) (6,86.2) (7,96.2) (8,92.8)};
\addlegendentry{System~1, DECISION}
\addplot[blue!70!black, thick, mark=square*, mark size=1.3pt] coordinates {(0,62.5) (1,72.5) (2,65.1) (3,79.2) (4,80.2) (5,85.4) (6,88.8) (7,97.4) (8,94.0)};
\addlegendentry{Opus + System~1, DECISION}
\addplot[black, thick, dashed, mark=o, mark size=1.3pt] coordinates {(0,23.8) (1,26.2) (2,33.9) (3,47.0) (4,73.3) (5,78.9) (6,89.5) (7,92.7) (8,88.2)};
\addlegendentry{System~1, BIRD dev}
\addplot[blue!70!black, thick, dashed, mark=square, mark size=1.3pt] coordinates {(0,86.9) (1,75.5) (2,75.5) (3,77.2) (4,88.7) (5,91.4) (6,91.7) (7,95.5) (8,91.0)};
\addlegendentry{Opus + System~1, BIRD dev}
\end{axis}
\end{tikzpicture}
\caption{Accuracy by System~1's validator margin (gap between its top two result groups; ``none'' = fewer than two groups, including every out-of-grammar question). System~1's accuracy rises with its margin on both splits, so the margin transfers. The agent's gain does not: above a margin of 2.5, Opus with candidates adds 1.2--2.6 points per band on DECISION (796 questions, 77--152 per band) but 12.5--15.4 in the two bands from 2.5 to 4.1 on BIRD dev and 2.2--2.8 above them (1,534 questions, 84--237 per band), which is why a threshold chosen on DECISION loses more on dev. Original gold.}
\label{fig:bands}
\end{figure}
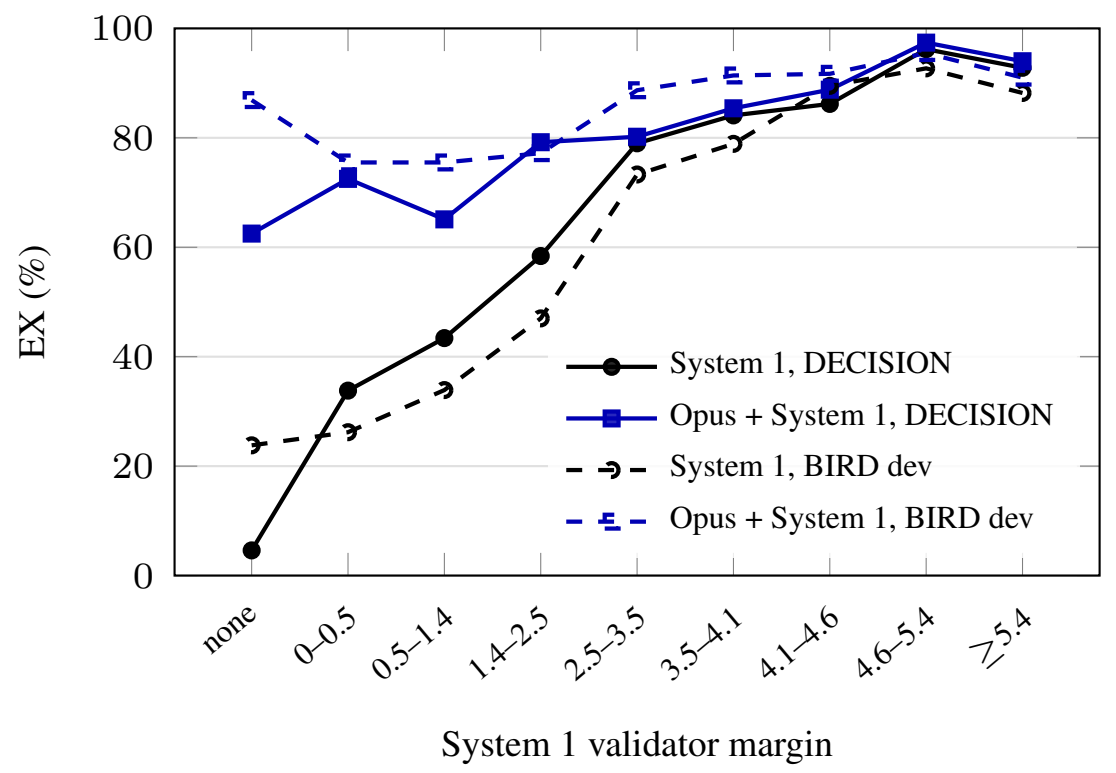

%% file: sections/app_tests.tex
\section{Held-out tests, probes and tuning}
\label{app:tests}
Every experiment here except the exploratory search at the end was registered, with endpoints and decision rules, before any agent ran on its data (Appendix~\ref{app:prereg}), and every agent transcript was scanned for access outside its sandbox.

\subsection{External test}
Against Opus with candidates, total time per question (summed agent time plus System~1's GPU time) fell by 12.3\%, 23.6\% and 21.7\% in the three RESERVED runs (19.3\% pooled; 2.90 against 3.59\,s per question on average) and by 17.1\% [12.5, 21.4] on the external test. Against a prompt-only control with empty candidate lists, Opus with candidates scored $-0.9$, $+1.7$ and $+1.4$ points in the three RESERVED runs (all $p \ge 0.5$), so the candidates add no clear accuracy.
KaggleDBQA's test split (185 questions, 8 databases) and Spider-Realistic (508 questions, 19 databases) were run blind: System~1 saw questions and schemas without SQL, and the agents only the databases, column descriptions and the tool. The registered claim pooled both sets (27 database clusters). On the registered clean subset of 130 KaggleDBQA questions, without two databases that overlap System~1's training data, the saving against Opus with candidates was 18.0\% [3.6, 32.9]; per set it was 25.3\% on KaggleDBQA and 34.6\% on Spider-Realistic. Simulated with the candidate arm's answers at the same call rate, routing by the margin scored 85.57 against 81.82 for vote agreement, 80.81 for question length and 78.89 at random.

\subsection{Memorization probes and renamed schemas}
\label{app:probes}
Probe agents had no database and no tool. In the schema-free probe each got the question and database name and wrote one SQLite query, scored by execution and by verbatim match after normalizing whitespace, quotes, case and backticks. In the completion probe each question was cut after half its words and completed with or without the dataset's name, in separate agents; guided minus general ROUGE-L was $+0.11$ to $+0.53$ on the four sets ($p<0.001$ each). For the renamed-schema test, a non-Claude model (DeepSeek V4.1 Flash) renamed every table and column of the 8 RESERVED databases with no shared word of three or more letters; gold and candidates were rewritten with a SQL parser and kept only if they returned the original result (all 177 gold queries passed), and questions and evidence were reworded with every literal kept. The routed agent ran unchanged.

\subsection{Fresh unseen set}
\label{app:fresh}
Eight SQLite databases were built, by Claude subagents doing data plumbing only, from public data dated 1 July 2026 or later (small reference tables excepted), each with at least three related tables and declared foreign keys: FDA enforcement reports (openFDA, CC0), USGS earthquakes and NWS alerts (US public domain), HM Land Registry price-paid data (OGL v3.0: ``Contains HM Land Registry data \copyright{} Crown copyright and database right 2021. This data is licensed under the Open Government License v3.0.''), the 2026 World Cup knockout matches (openfootball, CC0), Chicago crimes and traffic crashes (City of Chicago Data Terms of Use) and Formula~1 races (Jolpica-F1, CC BY-NC-SA 4.0). An MLB database was excluded because its terms forbid redistribution. DeepSeek V4.1 Flash wrote 40 questions per database with evidence and gold SQL that had to execute, be deterministic and return 1 to 1,000 rows; GLM~5.3 judged each item and the first 30 per database were kept. By an amendment recorded before any model ran, independent LLM agents (model not recorded; possibly Claude) wrote their own query before seeing the gold and judged all 240 items; 13 items with a majority of ``wrong'' verdicts were removed, leaving 227, and an author checked 20 by hand and found all correct. Counting the 13 removed items as misses gives 92.1\% for Opus alone and 89.2\% for the cascade. The questions often name columns or codes, so they are easier than BIRD's.

\subsection{Raising accuracy at the same cost}
\label{app:tuning}
\begin{table}[t]
\centering\small
\setlength{\tabcolsep}{2.5pt}
\begin{tabular}{@{}lccc@{}}
\toprule
Variant (routed questions) & EX & $-$ V0 & Cost vs V0 \\
\midrule
\multicolumn{4}{l}{Development half, renamed (92 questions; V0 76.09)} \\
Answer first & 76.09 & 0.00 & 1.09 \\
Answer first, top 3 cands & 75.00 & $-$1.09 & 0.92 \\
Second attempt, top 3 & 76.09 & 0.00 & 0.89 \\
+ independent 2nd agent, vote & 72.83 & $-$3.26 & 1.71 \\
\midrule
\multicolumn{4}{l}{Confirmation half, renamed (85 questions; V0 69.41)} \\
Second attempt, top 3 & 63.53 & $-$5.88 & 1.14 \\
\bottomrule
\end{tabular}
\caption{Attempts to raise the routed agent's accuracy at the same cost. V0: the cascade's routed arm. Cost: agent dollars per question relative to V0 on the same questions.}
\label{tab:tuning}
\end{table}
We registered variants of the routed agent and tested them on RESERVED's renamed copies, split by database into a development and a confirmation half (Table~\ref{tab:tuning}); none gained the required 2 points, and the best combination, run end to end on all of RESERVED, scored 80.17\% against the cascade's 80.46\% at a saving of only 19.1\%, so it was not adopted. While scoring them we found and fixed a bug that had costed the baseline over all routed questions but each variant over its half, which had made the top-3 variant look 20\% cheaper instead of 11\%. An exploratory search with 55 agents on DECISION only (\$2.98 of GPU) found no routing change worth more than 0.63 points.

\subsection{Opus through the API}
\label{app:plainapi}
The API agent (\texttt{experiments/plain\_api/}) calls Opus~5.5 through the Messages API with one conversation per question: the API-agent prompt of Appendix~\ref{app:prompts}, the schema output of the SQL tool in the user message, the read-only SQL tool with the eight-execution limit enforced in code, a submit tool, at most 14 turns, default sampling, no extended thinking, and prompt caching on the system prompt and the schema, so a database's schema is written to the cache once. Costs are its logged usage at the list prices of Section~\ref{sec:setup}.
The first registered API test on RESERVED kept the Claude Code design, in which each routed question also shows System~1's top eight candidate groups. Through the API those candidates add about 1,250 uncached input tokens and a third more output per routed question (\$0.0223 against \$0.0122 without them) and no accuracy (86.21\% against 86.49\% for Opus alone), so that cascade cost 18.3\% [2.3, 31.3] more than Opus alone, at 84.20\% accuracy; inside Claude Code the same difference was hidden by the fixed session cost. Forwarding only the question was then registered for the fresh set before any API call there (Section~\ref{sec:s2results}). The fresh run cost \$2.47 and the RESERVED run \$9.62.

\subsection{Registration record}
\label{app:prereg}
The registrations were private files in our development repository, each hashed (SHA-256) when written, with run records and amendments appended below the registered text; every stored hash equals that of the file's opening section except the external test's, whose placeholder for run details was filled in place, so its hash matches its first committed version. Table~\ref{tab:prereg} gives the commit times, which are our own records rather than an external timestamp; RESERVED run~1's registration and the cross-agent predictions were first committed together with their results. Future registrations will use an external registry.

\begin{table}[h]
\centering\small
\setlength{\tabcolsep}{3pt}
\resizebox{\columnwidth}{!}{\begin{tabular}{@{}lcc@{}}
\toprule
Experiment & Registered & Result \\
\midrule
RESERVED run 1 & \multicolumn{2}{c}{both 20th 21:45} \\
RESERVED runs 2--3, external & 20th 22:11 & 22:37; 21st 00:04 \\
Memorization probe & 21st 00:12 & 00:19 \\
Mini-Dev probe & 21st 08:54 & 09:00 \\
Renamed schemas & 21st 11:02 & 11:21 \\
Fresh set & 21st 14:29 & 17:55 \\
Tuning rounds 1--2 & 21st 18:42, 18:54 & 18:47, 18:57 \\
Final configuration C$^\ast$ & 21st 19:18 & 19:27 \\
API, RESERVED & 26th 12:22 & 12:44 \\
API, fresh set (question only) & 26th 13:15 & 13:20 \\
\bottomrule
\end{tabular}}
\caption{Commit times of each registration and of its first result in the development repository.}
\label{tab:prereg}
\end{table}

%% file: sections/appendix.tex
\section{Development-set use and possible leakage}
\label{app:leak}
BIRD dev was a development set in earlier rounds; after October 8, 2026 nothing was fitted or tuned on it, and LuthorSQL's configuration was hashed before its one scoring run. Some evidence-repair rules of earlier systems were written after reading dev misses; we report those systems with a version rebuilt without them, and LuthorSQL does not use that repair. Two Spider-train databases overlap dev (\texttt{formula\_1}; \texttt{soccer\_1} with \texttt{european\_football\_2}); retraining without them changed DECISION accuracy by $-1.01$ points ($p=0.42$), and on the other nine dev databases LuthorSQL scores 60.52\%. OmniSQL's fine-tuning data contains all of BIRD train, so DECISION and RESERVED are held out from our adapters only. Dev gold has been public since 2023, and Opus's answers alone match its gold text on 20.9\% of questions, so every agent setting was chosen on DECISION and dev agent scores are not treated as estimates for unseen questions.

\section{Compute and reproducibility}
\label{app:compute}
Provider billing records show 227.10 rented GPU-hours and \$464.46 from September 30 to October 13, 2026 (121.29 A100-hours, 96.38 H100-hours, 9.42 on other cards), covering every stage in Table~\ref{tab:history}, failed runs included; the later single-adapter work cost about \$26.6 by our run logs. R1 trained on one A100 80\,GB and VALHARD on one H100. The package (Python 3.12, PyTorch 2.11.0, Transformers 5.19.0, PEFT 0.21.2) loads the 7B backbone once per adapter (peak about 42\,GB) and answers 1.09 questions per second batched on one H100 PCIe. It was sent to BIRD on October 13, 2026, and BIRD's test result reached us on October 15. The DECISION and RESERVED ids are in the ancillary files with their SHA-256 hashes (of the ids sorted as strings and joined by newlines). The code, registrations, scorers, result files and prompts are at \url{https://github.com/IAMIbrahimmemon/luthorsql}; the adapters at \url{https://huggingface.co/IAMIbrahim/luthorsql-bird} (single adapter: \texttt{luthorsql-bird-single}) and the fresh set at \url{https://huggingface.co/datasets/IAMIbrahim/luthorsql-fresh-unseen}, for research use.

\section{Plan slots}
\label{app:slots}
Table~\ref{tab:slots} lists the slots of the plan grammar (v3).
\begin{table}[h]
\centering\small
\setlength{\tabcolsep}{4pt}
\begin{tabular}{@{}p{0.40\columnwidth}p{0.56\columnwidth}@{}}
\toprule
Slot & Values \\
\midrule
projections (1--4) & schema column or row count \\
aggregation & none, count, count distinct, sum, avg, min, max \\
expr.\ template & 533 templates (422 over columns, 111 over projections) \\
numeric constant & closed list (71) \\
conditional aggregate & condition on projection 1, 2 or both \\
filters (0--4) & column, operator, literal pointer \\
subquery filter & one inner plan (one projection) \\
group by (0--2), order by & columns, direction \\
HAVING (0--1) & aggregate, column, operator, literal \\
limit, distinct & small choices \\
date part, ratio, OR & choices \\
time grain & day, week, month, quarter, year \\
set operation & union, intersect, except, with a second plan \\
derived table (\texttt{wrap}) & aggregate over inner group-by \\
required tables (0--2) & tables joined without a used column \\
join hints & preferred bridge tables / keys \\
\bottomrule
\end{tabular}
\caption{Slots of the plan grammar (v3). Ranges give how many of each slot a plan can hold.}
\label{tab:slots}
\end{table}

\section{System~2 prompt and tool}
\label{app:prompts}
The routed agent received the prompt below, one per chunk of up to 25 questions, with \texttt{CHUNK} and \texttt{OUT} replaced by file paths. The agent-alone prompt, an example candidate list and the plain-API loop for the open-weight agents are in the code release. The tool is a read-only SQLite runner: \texttt{-{}-schema} prints the CREATE statements, three sample rows per table and BIRD's column descriptions, and a query prints the column names and at most 20 rows.

{\scriptsize
\begin{lstlisting}
You are answering BIRD text-to-SQL questions (SQLite) together with a fast text-to-SQL system. For each question that system already proposed candidate queries; they are grouped by identical execution result and listed best first by its own confidence (n_of_32 = how many of its 32 samples gave that result; "result" = a short preview of the rows). Some questions have no candidates.

Input: CHUNK (JSONL: question_id, db_id, question, evidence, candidates[{label, sql, result, n_of_32}]).
Output: append one JSON line per question to OUT as soon as it is done:
  {"question_id": <exactly as given>, "pick": "<label>"}            if one candidate answers the question correctly, or
  {"question_id": <exactly as given>, "pick": "NONE", "sql": "<your own SQLite query>"}   if none is correct or there are none.
Exactly one line per question.

Tool (the ONLY thing you may use besides reading CHUNK and writing OUT):
  python3 TOOLS/sql.py <db_id> --schema        # tables, sample rows, column descriptions
  python3 TOOLS/sql.py <db_id> "<SQL>"         # runs read-only SQL, prints up to 20 rows
Get each DB's schema only if you need it, and reuse it for later questions on the same DB.

Rules (strict -- this is a benchmark):
- Do NOT read any other file or directory (no repo files, no gold SQL, no other chunks/answers, no web).
- At most 8 SQL executions per question. Use the candidates to save work: when one is clearly right, pick it without extra exploration; write your own SQL only when you are confident every candidate is wrong.
- Judge correctness as the benchmark does: the result set must equal the intended answer (row order ignored). Use the evidence: it defines terms and formulas as the benchmark intends. Exactly the asked columns, nothing extra.
- If you write any helper script, put it ONLY in OUT's dir + "/../work/<chunk name>/"; never create or edit shared files (other agents run in parallel).
Final message: "done N/N, picked K, own M".
\end{lstlisting}}

\section{Further related work}
\label{app:related}
\paragraph{Autoregressive text-to-SQL.}
Early systems used grammar-constrained and execution-guided decoding \citep{scholak2021picard,wang2018executionguided}, alongside encoders that link question words to schema items \citep{wang2020ratsql,li2023resdsql}; LLM systems later added decomposition, self-correction, multi-agent pipelines and retrieval-based schema linking \citep{pourreza2023din,talaei2024chess,wang2025macsql}, and benchmarks such as Spider~2.0 now pose enterprise-scale workflows for such agents \citep{lei2025spider2}.
Several high-scoring open models build on large SQL corpora: CodeS \citep{li2024codes} pretrains open models for SQL, OmniSQL \citep{li2025omnisql} fine-tunes 7B to 32B models on millions of synthetic examples, and Arctic-Text2SQL-R1 \citep{yao2025arctic} trains with execution rewards starting from OmniSQL checkpoints.
We use OmniSQL-7B as our backbone but never decode text from it.

\paragraph{Slot filling and joins.} Before multi-table benchmarks, SQLova, X-SQL and HydraNet filled WikiSQL's sketch slots from pretrained encoders \citep{hwang2019sqlova,he2019xsql,lyu2020hydranet}. RYANSQL is the closest multi-table precedent: it splits nested SQL into non-nested SELECT statements and fills each with sketch-based slot filling on Spider \citep{choi2021ryansql}; our grammar is flatter (no nesting) but fills its slots from one pass of an LLM backbone and returns $k$-best plans. Our compiler's join inference over the foreign-key graph follows natural-language interfaces that connect the tables a query mentions with a Steiner tree over the schema or ontology graph, as in ATHENA and Templar \citep{saha2016athena,baik2019templar}. Pointing at stored values rather than generating them, as our filter slots do, appears in BRIDGE and ValueNet \citep{lin2020bridge,brunner2021valuenet}.

\paragraph{Selection and verification.} Contextual AI's BIRD system trains a reward model to pick among a generator's own candidates \citep{agrawal2025contextualsql}, as VALHARD trains on hard negatives drawn from our generator's candidate lists; it is described in a blog post and code release, not a paper.

\paragraph{Robustness and contamination.} \citet{rajkumar2022evaluating} gave Codex questions with no database information and argued qualitatively against memorization of Spider; our schema-free probe makes the same kind of test systematic and adds verbatim matching. Dr.Spider and Spider-Syn perturb schemas and questions \citep{chang2023drspider,gan2021spidersyn}, as our renamed-schema test does for the agent, and LiveSQLBench releases questions over newly built databases \citep{livesqlbench2025}, the idea behind our post-cutoff set.